\documentclass{article} 
\usepackage[margin=1.3in]{geometry}
\usepackage{graphicx}
\usepackage{epsfig}
\usepackage{amsmath,amssymb} 
\usepackage{color}
\usepackage{subfigure}
\usepackage{times}
\usepackage[ruled]{algorithm2e}
\usepackage{multirow}
\usepackage[T1]{fontenc}

\newcommand\T{\rule{0pt}{2ex}}       

\begin{document}

\title{An Adaptive Online HDP-HMM for \\ Segmentation and Classification of Sequential Data} 

\author{Ava Bargi, Richard Yi Da Xu, Massimo Piccardi \\ Faculty of Engineering and IT, University of Technology, Sydney \\ PO Box 123 Broadway NSW 2007 Australia\\ \tt\small {Ava.Bargi,YiDa.Xu,Massimo.Piccardi}@uts.edu.au }

\maketitle

\begin{abstract}
In the recent years, the desire and need to understand sequential data has been increasing, with particular interest in sequential contexts such as patient monitoring, understanding daily activities, video surveillance, stock market and the like. Along with the constant flow of data, it is critical to classify and segment the observations on-the-fly, without being limited to a rigid number of classes. In addition, the model needs to be capable of updating its parameters to comply with possible evolutions. This interesting problem, however, is not adequately addressed in the literature since many studies focus on offline classification over a pre-defined class set.
In this paper, we propose a principled solution to this gap by introducing an adaptive online system based on Markov switching models with hierarchical Dirichlet process priors. This infinite {\it adaptive online} approach is capable of segmenting and classifying the sequential data over unlimited number of classes, while meeting the memory and delay constraints of streaming contexts. The model is further enhanced by introducing a {\it `learning rate'}, responsible for balancing the extent to which the model sustains its previous learning (parameters) or adapts to the new streaming observations. Experimental results on several variants of stationary and evolving synthetic data and two video datasets, TUM Assistive Kitchen and collated Weizmann, show remarkable performance in segmentation and classification, particularly for evolutionary sequences with changing distributions and/or containing new, unseen classes.
\end{abstract}


\section{Introduction and related work}
The joint problem of time segmentation and recognition of sequential data into meaningful sub-sequences has attracted significant research in a variety of domains. The ability to automatically segment and classify data is a core technology for applications like speaker diarisation, finance, activity understanding, multimedia annotation and human-computer interaction. To date, the main proposed solutions have included sliding windows~\cite{bunke2007graph}, the hidden Markov model (HMM)~\cite{Yamato1992}, conditional random fields ~\cite{SminchisescuICCV2005} \cite{Vail2007}, and structural SVM~\cite{HoaiLD11}, covering the spectrum of generative, discriminative and maximum-margin dynamic classifiers. Along with advancements in learning and inference, research has witnessed increasingly realistic datasets which are bridging the gap between lab and real applications \cite{TRECVID2011} \cite{tenorth09dataset}. 

Nevertheless, important challenges such as model adaptation and dynamic class sets remain unresolved. We address both these limitations by an adaptive online model that can accommodate an unlimited (theoretically infinite) number of classes. In a nutshell, this is achieved by applying a Bayesian non-parametric model, the {\it hierarchical Dirichlet process} (HDP), as the prior for a hidden Markov model (a model known as HDP-HMM~\cite{TehJorBea2006}~\cite{beal2001infinite}), and exploiting an adaptive learning rate for model adaptation. The proposed model provides an adaptive online learning approach for joint segmentation and recognition of sequential data with incremental class sets and we refer to it as {\sc AdOn HDP-HMM} in the following. The model is i) {\it online}: can receive sequential data in batches and segment and recognise them on-the-fly; ii) {\it adaptive}: using a limited memory buffer, the model can tune its parameters in response to diverse observations from the existing classes, as well as instantiating new {\it unseen} classes. It continues learning throughout the entire life of its application; and is iii) {\it only-initially supervised}: the model uses a relatively short initial bootstrap of supervised training, but it adapts in a fully unsupervised manner during its operation. It is also considered as a one-pass process of streaming data, without revision. These constraints obviously make adaptation much more challenging, yet suiting the model to a large span of real-life problems. To improve adaptation in such an unsupervised learning scenario, we introduce the notion of {\it `learning rate'}, that tunes how biased the model is towards its previous learning (memory), versus adapting to the patterns conveyed by the new observations (adaptability). Experiments support the efficiency of utilising a learning rate, particularly in evolving scenarios.

The rest of this paper is organised as follows: in the rest of this Section we present the related literature and provide more clarification to the scope of this study. In Section~\ref{sec:HDP} we describe the hierarchical Dirichlet process and its temporal extension HDP-HMM. Section~\ref{sec:Online} presents the proposed online approach, expanding on the adaptive learning rate. Through the experiments and discussions in Section~\ref{sec:Experiments}, we evaluate and compare the proposed variants with existing benchmarks, and conclude in Section~\ref{sec:conclusion}.


\subsection{Related work}
\label{sec:literature}

Amongst the many paradigms available for class modelling, hierarchical Bayesian modelling and, in particular, the hierarchical Dirichlet process (HDP)~\cite{TehJorBea2006} offer a principled way to infer an arbitrary number of classes from a set of samples via a hierarchy of prior distributions. The hierarchical Dirichlet process (HDP) is a Bayesian nonparametric technique estimating the joint posterior distribution of a set of latent classes and a set of parameters, typically by Gibbs sampling ~\cite{Geman21984} or variational inference~\cite{TehKurWel2008}. It has been used for a variety of applications, including the modelling of sequential data, by integrating HDP priors into state-space models such as HMM. In the resulting HDP-HMM~\cite{TehJorBea2006}~\cite{beal2001infinite}, the classes correspond with the discrete states of a Markov chain and the data are explained by a state-conditional observation model. Given a set of samples, classification is performed by state decoding, while allowing the number of states to dynamically grow or shrink. The hierarchical Dirichlet process is finding increasing application in domains as varied as bio-informatics, speaker diarization, vision and others for problems of joint segmentation and classification (see~\cite{ZanottoICCVW13}~\cite{FoxTSP2011}~\cite{ZhangICCVW13} for some recent references).

Most of the segmentation and recognition studies in the literature follow an offline approach, where the entire data set is presented at once during the learning stage~\cite{TRECVID2011} \cite{tenorth09dataset}. Such systems obviously do not suit the needs of streaming data which are ubiquitous in today's applications. In response to this increasing demand for online systems, many studies are dedicated to this topic. However, the term \textit{online} has been given a variety of meanings in different contexts. Our interpretation is {\it sequential processing of temporal data in mini-batches}, inspired by recursive Bayesian estimation~\cite{Sorenson1971465} and further elaborated throughout this paper. This interpretation is distinct from that of other studies in the literature where online refers to a closed dataset that is processed incrementally and possibly repeatedly, such as Bayesian online nonparametrics \cite{OnlineLDA} \cite{OLDC_SSPR2008}, stochastic optimisation methods \cite{Adagrad2011} \cite{peskyLR_Arxiv2012}, formal bounds for online learning \cite{LampertICML2014}, all based upon the foundations laid by seminal works such as~\cite{Cesa-Bianchi}~\cite{Bottou2004}.

 
Despite that almost all the proposed approaches consider closed, pre-defined sets of classes, in scenarios like long-term learning or monitoring the number of classes is not precisely predictable. Additionally, as more data stream in, the known classes may change in parameters due to observing a more comprehensive sample or a natural evolution over time. In either case, models are expected to update parameters of the known classes and add new classes to their vocabulary once they appear. Unsupervised adaptation can be very challenging in non-stationary domains, where adaptation and drift\footnote{Defined as an undesirable deviation from the ideal model.} are hardly distinguishable. To our knowledge, a frequent assumption in online studies is to avail of periodic or ad-hoc feedback from the user (active learning \cite{LoyGong2012} \cite{Cesa-Bianchi} \cite{peskyLR_Arxiv2012}). This feedback allows the model to evaluate the {\it regret} and re-dress possible drifts and misclassifications. However, such information is hard or costly to obtain in many real application domains.

In the absence of expert feedback, we elaborate more on the {\it learning rate} as a dynamic lever for balancing adaptability (section \ref{sec:adaptive}). Most previous studies approach this problem by assigning constant weights to prior learning and the likelihood of the current data. However, in more complex problems the choice of the learning rate is highly dependent on the data dynamics and the application domain. Some online studies propose adaptive learning rates via exponential decay \cite{decayLR1951}, and, more recently, regret-based adaptations of the learning rate (i.e., the step size of gradient descent) \cite{Adagrad2011} \cite{OLDC_SSPR2008} \cite{peskyLR_Arxiv2012}. However, such adaptation strategies are only suitable for finite training sets. In our solution, we introduce a novel learning rate that constantly adapts to the statistics of the streaming data, without revision or supervision. For stationary problems where the parameters only slightly change, the learning rate tunes itself to rely more on prior memory. Conversely, under evolving distributions the dynamics of data and their modes can significantly vary, calling for a more adaptive model with less inertia to the past. Adding to the complexity, many real-life problems require a mixture of both, i.e. a continuous spectrum for the learning rate to follow more or less tightly the dynamics of observations at each point in time. In this work, we tackle this problem by a posterior estimation of the learning rate separately for each parameter in the model - thereby, allowing each parameter to dynamically determine its adaptability in each batch.



\section{The hierarchical Dirichlet process}
\label{sec:HDP}

A Dirichlet process, $DP(\gamma, H)$, is a generative model that can be thought of as a distribution over discrete distributions with countably infinite categories. It is controlled by a scalar parameter, $\gamma$, known as the concentration parameter, and a base measure, $H$, over a measurable space $\theta$. A sample $G_0$ from a Dirichlet process is a distribution over $\theta$ differing from zero at only a countably infinite number of locations or atoms, $\theta_k, k=1 \ldots K$:

\begin{equation}
\label{eq:DP}
\begin{split}
& G_0 \sim DP(\gamma, H):\\
& G_0 = \sum_{k=1}^K  \beta_{k} \delta (\theta - \theta_k),  \hspace{0.5 cm}   {K \to \infty}\\
& \theta_k \sim H, \hspace{0.5 cm} \mathbf{\beta} \sim GEM(\gamma)
\end{split}
\end{equation}

\begin{figure}
\centering
\includegraphics[scale=0.2, height=4cm]{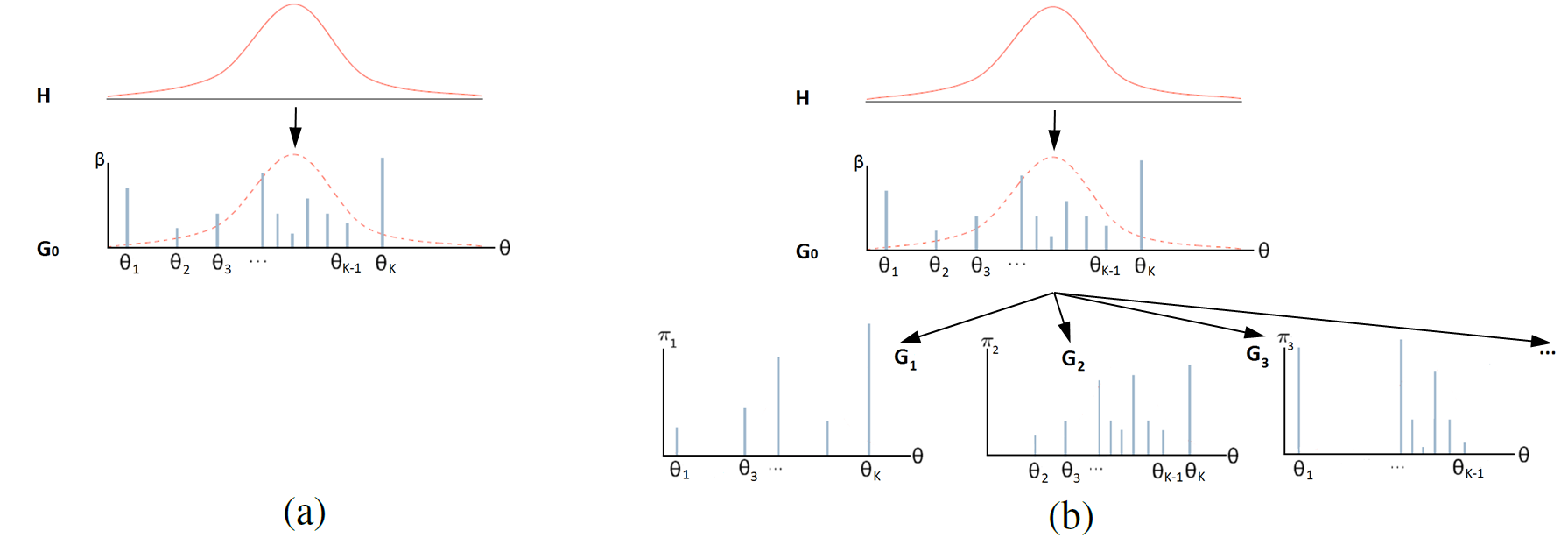}
\caption{The Dirichlet Process (a) and Hierarchical Dirichlet Process (b) construction; the parameter space has been simplified to one dimension for the sake of visualisation.} \label{fig:DPHDP}
\end{figure}

The discrete set of locations is obtained by repeatedly sampling the base measure, while the weight for each location, $\beta_k, k=1 \ldots K$, is established by a \emph{stick-breaking process}, noted as GEM($\gamma$) (named after Griffiths, Engen and McCloskey)~\cite{Sethuraman}. We refer to the weight vector simply as $\beta$. A \emph{hierarchical Dirichlet process} (HDP) consists of (at least) two layers of Dirichlet processes, obtained with a similar construction:

\begin{equation}
\label{eq:HDP}
\begin{split}
& G_j \sim HDP(\gamma, \alpha, H):\\[2\jot]
& G_0 \sim DP(\gamma, H)\\
& G_j = \sum_{k=1}^K  \pi_{jk} \delta (\theta - \theta_k)  \hspace{0.5 cm}   {K \to \infty}\\
& \theta_k \sim H, \hspace{0.5 cm} \pi_{j} \sim DP(\alpha, \beta), \hspace{0.5 cm} \mathbf{\beta} \sim GEM(\gamma)
\end{split}
\end{equation}

where $\gamma$ and $\alpha$ are the concentration parameters of the top-level an lower-level Dirichlet processes, respectively. Since $G_0$ is discrete, the various $G_j$ $(j = 1 \ldots J)$, are also discrete and sampled from the elements of $G_0$ (Figure \ref{fig:DPHDP}). 

In practical applications, the continuous space of distribution $H$ is taken to be the parameter space for a data likelihood, as in $y \sim f(y|\theta): \hspace{0.2 cm} \theta \sim H$. Likelihood $f(y|\theta)$ could be, for instance, a Gaussian distribution of mean parameters $\theta$ sampled from a Normal-Inverse-Wishart (NIW) distribution. Given the generative model of the HDP, the joint distribution of data and parameters factorises as $f(y|\theta) G_j(\theta)$. Typically, multiple $G_j$ are sampled to model data belonging to different groups. Yet, the hierarchical structure of the HDP makes all the $G_j$ usefully share distributional properties. Examples can be as diverse as words in a collection of books or genetic markers across different populations. 

\subsection{The HDP-HMM}

\begin{figure}
\centering
\includegraphics[scale=0.17]{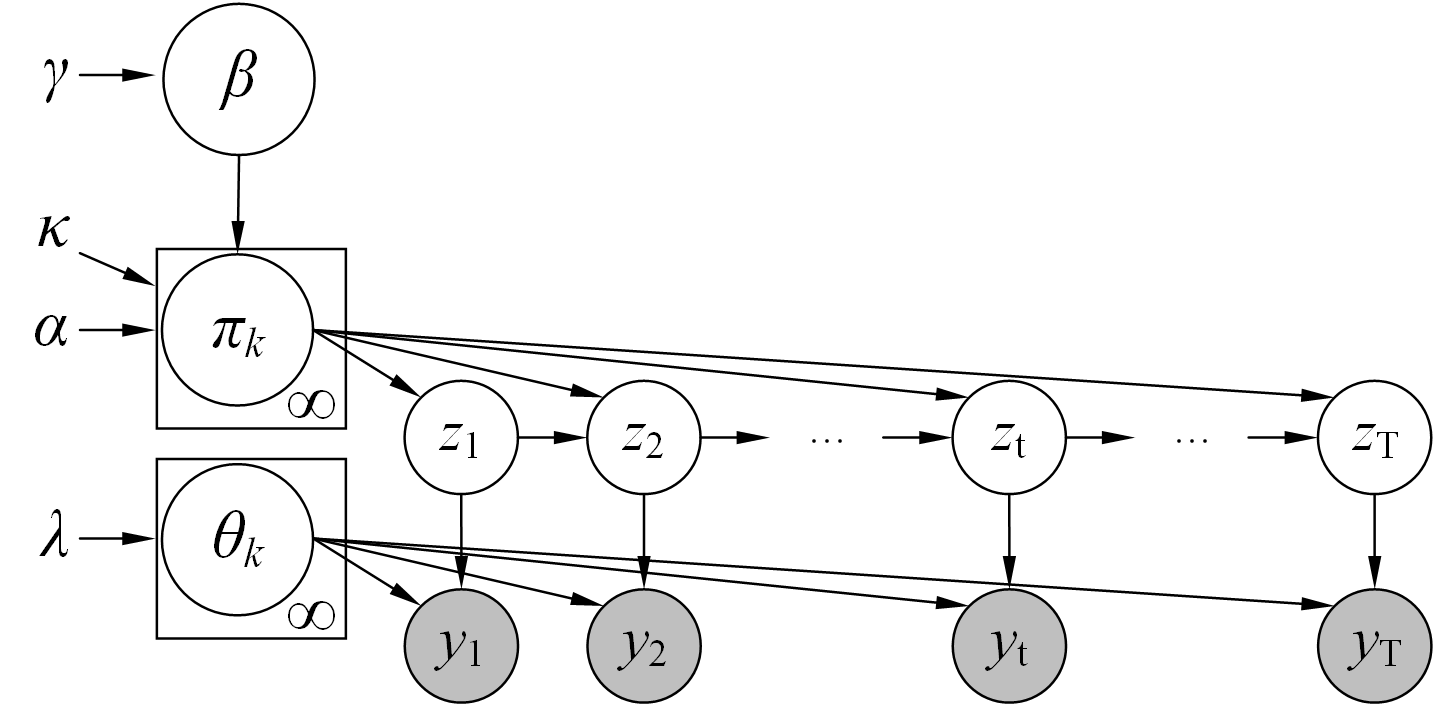}
\caption{HDP-HMM graphical model. The box notation is used to show replication.} \label{fig:HDPHMM}
\end{figure}

The HDP has also been used as prior distribution for the parameters of switching models such as the hidden Markov model~\cite{TehJorBea2006}~\cite{FoxTSP2011}. When applied to a Markov chain, $z_{1:T}$, $p(z_{1:T})=p(z_{1})\prod_{t=2}^T p(z_{t}|z_{t-1})$, the HDP changes its interpretation significantly (Figure~\ref{fig:HDPHMM}). In this case, each $\pi_j = \left\{\pi_{jk} \right\}$, ${k=1 \ldots K}$, is used as one row of the Markov chain's transition matrix, representing the probability of transitioning from state $j$ in the previous time-step to any other states in the current time-step, $p(z_t|z_{t-1}=j$). Thanks to the properties of HDP, new states will be created when the data are not adequately explained by the current set of states. In contrast to the conventional HDP, the index of the group, $j$, of each observation is usually not known explicitly anymore, but it is instead inferred in sequential order from the chain. Therefore, in the case of the HDP-HMM $z_t \sim p(z_{t}|z_{t-1} = j) = \pi_{j}, \hspace{3mm} y_t \sim f(y_t|\theta_{z_t})$. As a consequence, in the HDP-HMM the number of groups ($J$) and the number of indices in each $\pi_j$ ($K$) coincide. Adding the HDP as prior caters for arbitrary number of states, or activity classes~\cite{FoxTSP2011}.

It is worth adding that a reported limitation of HDP-HMM is the tendency to over-segment due to its unbounded number of classes \cite{Fox:2007}. Fox \textit{et al}. have proposed adding a {\it `sticky'} prior ($\kappa$) to the transition matrix to emulate an inertia towards changing states, illustrated in Figure \ref{fig:HDPHMM} \cite{FoxICML08}. We utilise the {\it sticky} prior in this study, yet denoting it as HDP-HMM for brevity.

\subsection{Inference and Learning}

Inference and learning are typically performed simultaneously in the HDP and its extensions by estimating the joint posterior distribution of the indicator variables, parameters, hidden variables and hyper-priors conditioned on the observations. Deriving such an extensive joint posterior is analytically intractable, hence mainly inferred using Gibbs sampling or variational inference. Gibbs sampling is a simple yet effective method capable of estimating complex posteriors with significant accuracy, yet it can converge slowly or permanently remain in a local minima ({\it poor mixing}). Variational inference is usually faster to compute, however it requires prior derivation of analytical approximations and can suffer from low accuracy due to the approximation. Unlike the negative presumption about Gibbs efficiency, we will show how a brief initial supervised learning can result in rapid convergence to accurate distributions.

Having inferred the class indicators, $z_{1:T}$, we proceed with translating the indices into meaningful classes. In unsupervised learning, the correspondence between the ground-truth classes of data and the labels assigned by the classification algorithm may not be obvious. In the case of the HDP, this problem is exacerbated by the fact that the number of classes is undetermined. Therefore, to re-establish the best possible one-to-one correspondence, the Hamming distance between ground-truth and assigned labels is minimised by a greedy algorithm, matching labels in decreasing frequency order.


\section{The Adaptive Online HDP-HMM}
\label{sec:Online}

\begin{figure*}
\centering
\includegraphics[scale=0.19]{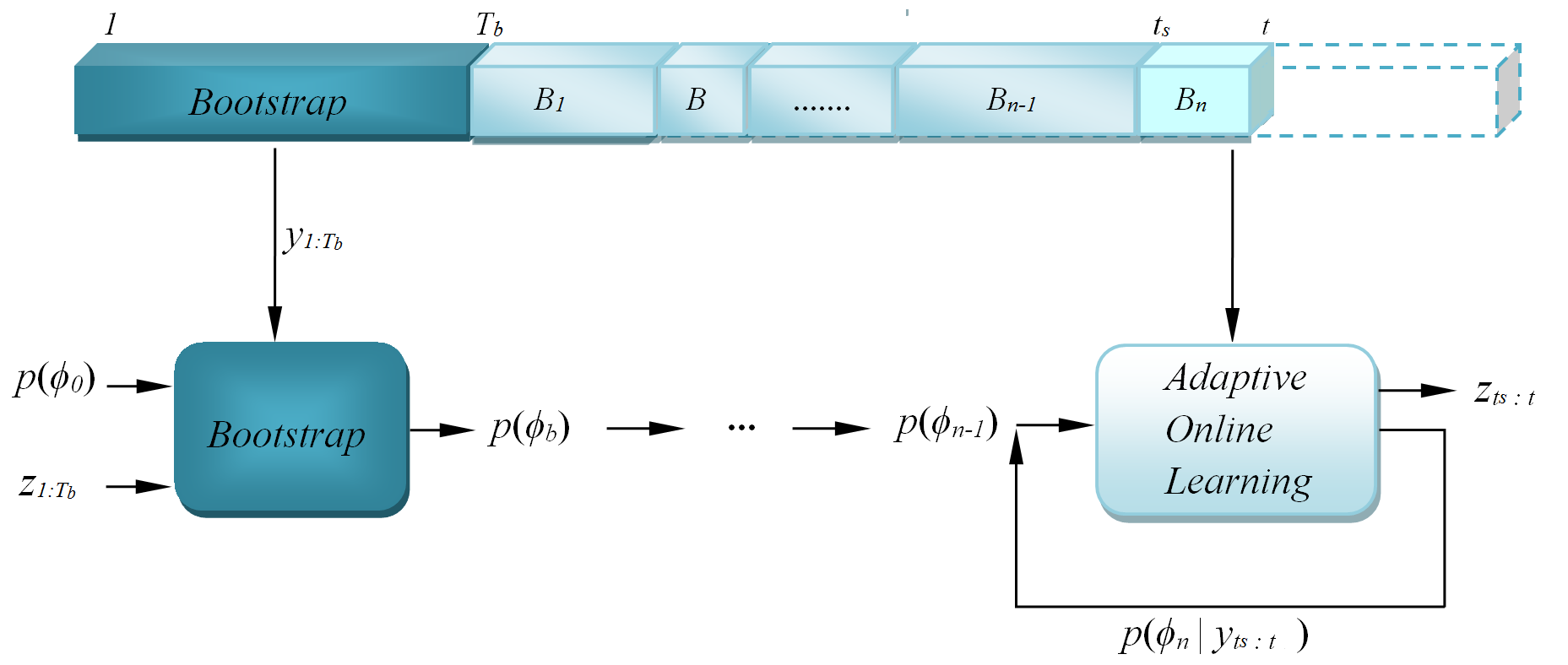}
\caption{Adaptive online learning flowchart: initialised by a supervised bootstrap, the learning continues unsupervised over streaming data split into batches. The figure shows a general case with batches of variable size. For simplicity we assume all batches to have the same size (see \cite{UsICPR14} for the variable alternative). The posterior $\phi$ in each batch is passed to the next batch as prior parameters.} \label{fig:flowchart}
\end{figure*}

The proposed \textsc{AdOn HDP-HMM} uses a supervised initialisation (bootstrap) of $T_b$ frames, followed by the main unsupervised adaptive online inference (Figure~\ref{fig:flowchart}). The extent of the supervised phase varies with the application: in applications where annotation is easy, the bootstrap can be longer to provide a more comprehensive training, while in domains with costly annotation the bootstrap will be brief. In either case, during supervised learning, indicator variables $z_{1:T_b}$ are fixed to their ground-truth values, and the model's parameters are sampled for a given number of iterations to reach convergence. After conclusion of the bootstrap phase, the data are processed in successive batches, and the posterior probabilities of both indicator variables and parameters are estimated iteratively on each batch. 

\begin{figure*}
\vspace{-0.5cm}
\centering
\includegraphics[scale=0.19]{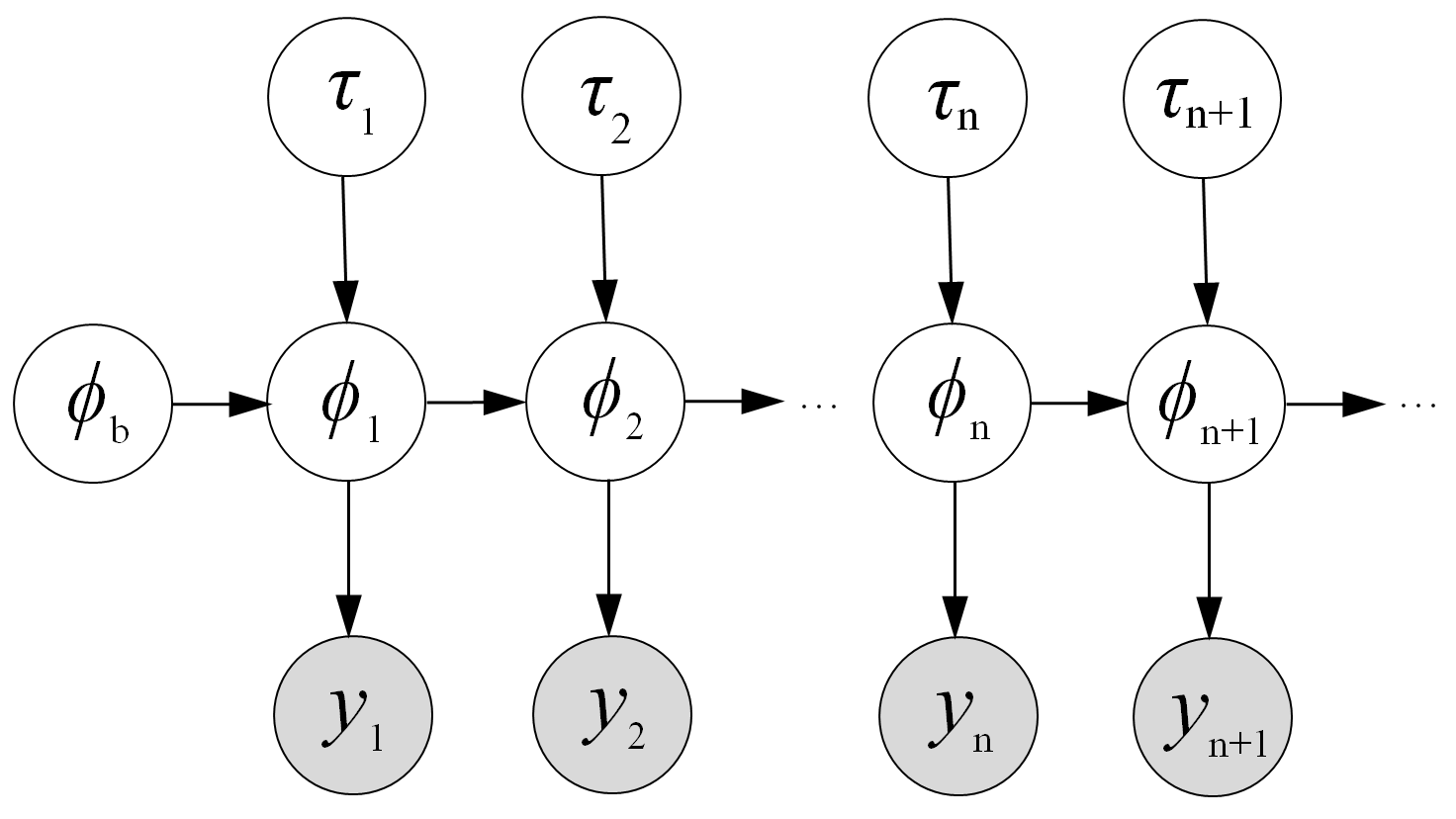}
\caption{Graphical model for the proposed adaptive online model. $\phi$ can be any of the parameters in Figure \ref{fig:HDPHMM} ($\theta, \pi, \beta$), $\tau$ is the respective learning rate (a positive continuous variable) and $n$ represents the batch number.} \label{fig:onlineGM}
\end{figure*}

Considering a generic stream of data, $y_{1:t}$, the posterior probability of the parameters can be written as $p(\phi|y_{1:t}) \propto f(y_{1:t}|\phi) \hspace{0.1 cm} p(\phi)$, where $\phi$ indicates the parameter vector of Figure \ref{fig:HDPHMM}. In the case of the HDP-HMM, the parameter vector is $\phi = \left\{ \theta,\pi,\beta \right\}$ where $\theta$ are the parameters of the emission densities, $\pi$ are the transition probabilities (and weights of the lower-level DPs), and $\beta$ are the weights of the higher-level DP. Further, since we assume normal densities, we have $\theta = \left\{ \mu, \Sigma \right\}$, with $\mu$ and $\Sigma$ the usual mean and covariance parameters. The online version leverages on posterior adaptation, using the posterior computed up to time $t$, as the prior for the next batch of data, $y_{t + 1:t + \Delta t}$:

\begin{equation}
\label{eq:conjugacy_online}
\begin{split}
p(\phi_{n+1}|y_{1:t + \Delta t}) & \propto f(y_{t + 1:t + \Delta t}|\phi_{n}, y_{1:t}) \hspace{0.1 cm} p(\phi_{n}|y_{1:t}) \\
& \approx f(y_{t + 1:t + \Delta t}|\phi_{n}) \hspace{0.1 cm} p(\phi_{n})
\end{split}
\end{equation}

where $n$ is the batch number (Figure \ref{fig:onlineGM}). Given that the updated posterior embeds the distributional properties of the observations up to the current time, observations $y_{1:t}$ in Equation \ref{eq:conjugacy_online} can be discarded after adaptation. It implies that the accumulated sufficient statistics of previous data are propagated parametrically and the non-parametric nature of the model is related to the inference method of the current data batch. With that, the model carries all the prior learning and infers new labels using a limited memory buffer. While this may come at a price of reduced accuracy, to our knowledge it is the only viable approach for unbounded streaming data. 
In contrast ~\cite{OnlineLDA} presents online inference for latent Dirichlet allocation, yet over an unbounded buffer. Our work extends that model to infinite class sets while meeting the finite memory requirements of sequential data processing.  

\subsection{Learning rate adaptation}
\label{sec:adaptive}

In the proposed adaptive system, a learning rate is applied over the prior and noted as $\tau$ in the following. In each batch, $\tau$ is responsible for setting the weight of prior learning on the model's parameters ($\theta, \pi, \beta$). In other words, our target is to balance the impact of the current observations with the previous learning accumulated along the previous batches. This can augment or weaken the posterior learning `inertia' in `adapting' to the current data (likelihood), as opposed to retaining `memory' (prior).

\begin{equation}
\begin{split}
p(\phi | y, \tau) \propto p(y|\phi) \boldsymbol{p (\phi)^{\tau}}\\
\end{split}
\label{eq:learningrate}
\end{equation}

It is worth noting that the length of the current batch compared to the number of past samples, plays a role in their relative influence on posterior parameters (see Appendix A for more details). Accordingly, $\tau$ can be articulated as a scaling factor to the number of `pseudo-observations' in the prior to balance with the respective number for current batch\footnote{For convenience, in this paper we have constrained all batches to be of the same length and explored the variable alternative in \cite{UsICPR14}.}.

For prior distributions belonging to the exponential family, this proposition does not violate Bayes' Theorem, thanks to the properties of canonical parameters. Accordingly, we use exponential family likelihoods and priors for easier integration of the learning rate into the model. Hereby, we focus on the prior in Equation \ref{eq:learningrate} (in bold font) and its hyperparameters, translating them into exponential family notations. The standard parameters, $\phi$, are converted into the corresponding canonical parameters, $\Theta$, and we make explicit their dependence on hyper-parameters, $\eta$:

\begin{equation}
\begin{split}
& p(\Theta|\eta)^{\tau} = p(\Theta|\eta, \tau) = f ( \eta)^\tau g(\Theta)^\tau \exp \left( \Theta^T \eta \right)^\tau\\
& \hspace{3cm} = f' ( \eta) \exp \left( \tau \Theta'^T \eta \right), \hspace{1cm} \Theta' = [ln(g(\Theta)); \Theta]\\
\end{split}
\label{eq:postTheta}
\end{equation}

Adding the learning rate ($\tau$) as an exponent to this prior does not alter the type of distribution. Rather, it updates the canonical parameters of the prior, ultimately affecting its weight in the resulting posterior. Please note that we only need to derive a proportional posterior for sampling purposes. Hence, the $\tau$ exponent on any term independent from $\Theta$ (such as $f(\eta)$) can be ignored thanks to the proportionality. The normalisation coefficient $g(\Theta)^\tau$ can be merged into the sufficient statistics, assuring that its $\tau$ exponent is absorbed into the scaled canonical parameter ($\tau \eta$). \footnote{$g(\Theta)^\tau \exp(\tau \Theta^T \eta) = \exp \left( \tau ln(g(\Theta)) + \tau \Theta^T \eta) \right) = \exp \left( \tau [ln(g(\Theta)); \Theta]^T[1; \eta] \right)$}


In general terms, the posterior distribution of $\tau$ given $\Theta$ in the presence of $N$ data samples in ${\bf Y}$ can be inferred as follows:

\begin{equation}
\begin{split}
p(\tau | \Theta, {\bf Y}, \eta) \propto p(\Theta | \tau, {\bf Y}, \eta) p(\tau) \\
\end{split}
\label{eq:taupost}
\end{equation}

In our case, $\Theta$ are the parameters of the HDP-HMM and their priors are a Normal-Inverse-Wishart distribution for $\mu$ and $\Sigma$ and the HDP for $\pi$ and $\beta$. Given that both the NIW distribution and the Dirichlet process are members of the exponential family, Equation \ref{eq:postphi} shows a unified way of inferring posterior parameters in canonical form \cite{bishop2006pattern}:

\begin{equation}
\label{eq:postphi}
\begin{split}
&p(\Theta | {\bf Y}, \tau^*, \eta^*) \propto p({\bf Y} | \Theta, \tau, \eta) p(\Theta | \eta, \tau)\\
&p(\Theta | {\bf Y}, \tau^*, \eta^*) \propto  \left[ \left( \prod_{n=1}^N  h(y_n) g(\Theta) \right) \exp \left( \Theta^T \sum_{n=1}^N u (y_n) \right) \right] \left[f ( \eta, \tau)g(\Theta)^{\tau} \exp \left( \tau \Theta^T \eta \right) \right] \\
&\text{\it \small removing the constants with respect to $\Theta$: }  \\
&p(\Theta | {\bf Y}, \tau^*, \eta^*) \propto g(\Theta)^{\tau + N} \exp \left( \Theta^T \left(   \sum_{n=1}^{N} u(y_n) + \tau \eta    \right)  \right)\\
& \tau^* = \tau + N, \hspace{2cm} \eta^* = \sum_{n=1}^{N} u(y_n) + \tau \eta \\
\end{split}
\end{equation}

In the following sub-sections, we present the prior distribution of each parameter under the learning rate, and the posterior distribution of the corresponding learning rate.

\subsubsection{Inference of covariance matrix $\Sigma$}
We infer $\mu$ and $\Sigma$ in the Normal-Inverse-Wishart prior by first sampling $\Sigma$ using an Inverse-Wishart (IW) distribution, thereby using $\Sigma$ to sample $\mu$ from a Normal distribution~\cite{emilythesis}. The learning rate for $\Sigma$ is noted as $\tau_{\Sigma}$ in the text. Yet, to avoid cluttering the notation in the equations, we simply note it as $\tau$ in Equation \ref{eq:Sigma}.

\begin{equation}
\label{eq:Sigma}
\begin{split}
&p (\mu, \Sigma)^\tau = p (\mu | \Sigma)^\tau p (\Sigma)^\tau: \\
&p(\Sigma) = IW (\Sigma| \Psi, \nu), \hspace{0.3cm} p (\mu | \Sigma) = \mathcal{N} (\mu | \mu, \frac{1}{\sigma} \Sigma)\\
\end{split}
\end{equation}

As mentioned earlier, the addition of a positive learning rate as exponent on the IW prior does not alter the type of distribution and can be merged into the hyper-parameters. Below, we convert the hyper-parameters $\phi_{IW} = \{ \Psi, \nu \}$ into the natural form ($\eta$) to show the impact of $\tau$ more clearly. Ultimately, they are converted back to standard form ($\phi$) to show the linear transformation caused by the learning rate.

\begin{equation}
\begin{split}
&\phi_{IW} = (\Psi, \nu) \rightarrow \eta_{IW} =\left(-\frac{1}{2} \Psi,  - \frac{\nu+p+1}{2} \right), \hspace{0.7cm} p= \text{\it \small number of dimensions}\\
&\eta'_{IW} =  \tau\eta_{IW} = \left(-\frac{\tau}{2} \Psi,  - \frac{\tau(\nu+p+1)}{2} \right) \rightarrow \phi'_{IW} = (\tau \Psi, \tau (\nu + p + 1) - p - 1) \\
\end{split}
\end{equation}


{\bf Inference of ${\bf \tau_{\Sigma}}$}

To sample $\tau_\Sigma$ from the posterior, ideally we would like to consider a conjugate prior that analytically derives the posterior hyper-parameters, given that of the prior and the sufficient statistics of the current data. A candidate conjugate prior for IW distribution is Gamma. However, the Inverse-Wishart is only conjugate to the Gamma as the prior for the scale parameter (or a scaling coefficient for the scale parameter, $\Psi$, in the multivariate cases). Hence, a Gamma cannot be used as a conjugate prior for deriving the posterior of $\tau_\Sigma$ in a maximum-a-posteriori solution (Appendix B presents the proof). 

Therefore, we utilise a maximum-likelihood solution to derive the posterior hyper-parameters for $\tau_\Sigma$. The posterior for $\tau_\Sigma$ is modeled using an Inverse-Gamma (IG) distribution, the univariate correspondent of the Inverse-Wishart. The samples of IG are positive real values, suitable for the scalar learning rate $\tau_\Sigma$. The distributions are displayed below.

\begin{equation}
\label{eq:IWvsIG}
\begin{split}
&IW(\Sigma|\Psi, \nu) = \frac{|\Psi|^\frac{\nu}{2}}{2^{\frac{\nu}{2}} \Gamma(\frac{\nu}{2})}  |\Sigma|^{-\frac{\nu + p + 1}{2}} \exp\left(-\frac{1}{2} tr(\Psi\Sigma^{-1}) \right) \\
&IW(\sigma|\psi, \nu) = \frac{\psi^\frac{\nu}{2}}{2^{\frac{\nu}{2}} \Gamma(\frac{\nu}{2})}  \sigma^{-\frac{\nu + 2}{2}} \exp\left( {-\frac{\psi}{2\sigma}} \right) \hspace{1cm} \text{\it \small univariate IW : p = 1}\\
&IG(\tau|\beta, \alpha) = \frac{\beta^{\alpha}}{\Gamma(\alpha)}  \tau^{-\alpha-1} \exp\left( {-\frac{\beta}{\tau}} \right) \hspace{1cm} \\
\end{split}
\end{equation}

Comparing the univariate IW and IG in Equation \ref{eq:IWvsIG}, we can derive the posterior parameters as:

\begin{equation}
\label{eq:IGvsIW}
\begin{split}
&IG(\tau|\beta^*, \alpha^*) \approx IW(\Sigma|\Psi, \nu) \hspace{1cm} \text{\it \small where } \beta^* = \frac{f(\Psi)}{2}, \alpha^* = \frac{\nu}{2}\\
\end{split}
\end{equation}

\begin{center}

\end{center}

As can be seen, the hyper-parameters in the Inverse-Wishart map to those of the Inverse-Gamma. The only issue is how to best map the $p \times p$ scale matrix ($\Psi$) into  scalar value $\beta^*$ through $f(\Psi)$. We propose to use the largest eigenvalue in $\Psi$ as the scale parameter in the Inverse-Gamma posterior. Approximating a covariance matrix via its first principal components is a meaningful and common approach \cite{Tipping99} \cite{Chiani201469}. Another choice for $f(\Psi)$ could be the determinant. The determinant is used as the scale associated with a square matrix since it is equal to the product of all its eigenvalues. While it gives a more thorough account of all the eigenvalues, it becomes unsuitable when the dimensionality is high and many of the eigenvalues are close or equal to zero. Moreover, calculating the determinant of a high-dimensional matrix is very costly in an online context. Therefore, the first approach is generally preferable.\\


So far, we have established a way to infer $\tau_\Sigma$ from a single IW distribution belonging to a single state. Considering the proposed model with infinite states, we need to merge the IW parameters for all classes to infer $\tau_\Sigma$. This is done through a weighted average of $\Psi_k, k = 1 \ldots K$, where the weights are the frequency of observations for each state, aka degrees of freedom in IW parameters ($\nu$).

\begin{equation}
\label{eq:IGvsIW}
\begin{split}
&IG(\tau|\beta^*, \alpha^*) \approx IW(\Sigma|\Psi, \nu) \hspace{1cm}  \beta^* = \frac{\sum_{k=1}^K{max(eig(\Psi_k)) . \nu_k}}{2 \sum_{k=1}^K{\nu_k}}, \alpha^* = \frac{\sum_{k=1}^K{\nu_k}}{2K}\\
\end{split}
\end{equation}
 
\subsubsection{Inference of mean $\mu$}
Having inferred $\Sigma$, the next step is to derive the multivariate mean, $\mu$, in the NIW prior. Let us consider a generic multivariate Normal distribution $N = (\mu |\mu_0, \frac{1}{\sigma}\Sigma)^T$ with known covariance. To observe the impact of the learning rate, we convert its parameters $\phi_{\tiny \mu} = (\mu_0, \frac{1}{\sigma}\Sigma)$ into the natural form and multiply them by the learning rate $\tau_\mu$, and ultimately revert them back to the standard format:

\begin{equation}
\begin{split}
&\phi_{\tiny \mu} = (\mu_0, \frac{1}{\sigma}\Sigma) \rightarrow \eta_{\small \mu} =\left(\frac{1}{\sigma} \Sigma ^ {-1} \mu_0,  \frac{1}{2\sigma} \Sigma^{-1} \right)^T \\
&\eta'_\mu = \tau \eta_\mu = \left(\frac{\tau}{\sigma} \Sigma ^ {-1} \mu_0,  \frac{\tau}{2\sigma} \Sigma^{-1} \right)^T
 \rightarrow  \phi'_{\tiny \mu} = \left( \mu_0, \frac{1}{\tau \sigma}\Sigma  \right) \\
&N(\mu|\eta'_{\small \mu}) = \mathcal{N} (\mu | \mu_0, \frac{1}{\tau \sigma}\Sigma  )\\
\label{eq:tau_mu}
\end{split}
\end{equation}

{\bf Inference of $ {\bf \tau_\mu}$}

Posterior sampling of $\tau_\mu$ is conducted with a similar approach to $\tau_\Sigma$, but using a Gamma conjugate prior. This time the Gamma prior is conjugate by definition, since its sample $\tau_\mu$ is utilised merely as a scaling coefficient for the covariance. The detailed proof of the conjugacy is provided in Appendix C.
Similarly to $\tau_\Sigma$, the weighted average of sufficient statistics across all classes is used to infer $\tau_\mu$.

\begin{equation}
\begin{split}
&G(\tau | \alpha^*, \beta^*) \propto \mathcal{N}(\mu| \mu_0,\frac{1}{\tau \sigma}\Sigma ) G(\tau | \alpha, \beta)\\
&\alpha^* = \alpha + 1/2 , \hspace{1cm} \beta^* = \beta + \frac{\sigma \sum_{k=1}^K{(\mu_k - \mu_{0k})^T \Sigma_k ^{-1} (\mu_k - \mu_{0k}) . \nu_k}}{2\sum_{k=1}^K{\nu_k}} \\
\end{split}
\end{equation}

\subsubsection{Inference of the HDP transition parameters}
Thus far we have discussed the adaptation of the learning rate for emission parameters. The other main set of parameters in our {\sc AdOn HDP-HMM} are the HDP's $\beta$ and $\pi$ parameters that jointly and hierarchically cater for the transition probabilities. The distributions of these parameters are shown in Equation~\ref{eq:inferDir}, where $m$ and $ n$ are HDP sufficient statistics representing the frequency of occurrence in each class:

\begin{equation}
\begin{split}
\label{eq:inferDir}
&{\bf \beta} \sim Dir(\gamma/L + m_{.1}, ..., \gamma/L + m_{.L})\\
& \pi_j \sim Dir(\alpha_1 \beta_1 + n_{j1}, ... , \alpha_j \beta_j + \kappa + n_{jj}, ..., \alpha_L \beta_L + n_{jL})\\
\end{split}
\end{equation}

Similarly to the previous parameters, we illustrate the impact of the learning rates, $\tau_\beta$ and $\tau_\pi$, on the hyper-parameters of the above Dirichlet distributions in {\it standard} form and infer the posterior samples for these learning rates:

\begin{equation}
\begin{split}
&\text{{\bf Inference of the learning rate for }} {\bf \beta} \\
&\phi_\beta = (\gamma/L + m_{.1}, ..., \gamma/L + m_{.L}) \rightarrow \eta_\beta = (\gamma/L + m_{.1} -1 , ..., \gamma/L + m_{.L} -1 )\\
&\eta'_\beta = \tau_\beta \eta_\beta = (\tau_\beta(\gamma/L + m_{.1}) -\tau_\beta , ..., \tau_\beta(\gamma/L + m_{.L}) -\tau_\beta )\\ \\
&\beta \sim Dir(\tau_\beta(\gamma/L + m_{.1} - 1) +1 , ..., \tau_\beta(\gamma/L + m_{.L} - 1) +1)\\ \\
&\text{{\bf Inference of the learning rate for }} {\bf \pi} \\
&\phi_\pi = (\alpha_1 \beta_1 + n_{j1}, ... , \alpha_j \beta_j + \kappa + n_{jj}, ..., \alpha_L \beta_L + n_{jL}))\\
&\rightarrow \eta'_\pi = (\tau_\pi(\alpha_1 \beta_1 + n_{j1}) -\tau_\pi, ... , \tau_\pi(\alpha_j \beta_j + \kappa + n_{jj}) -\tau_\pi, ..., \tau_\pi(\alpha_L \beta_L + n_{jL} )-\tau_\pi)\\ \\
&\pi_j \sim Dir(\tau_\pi(\alpha_1 \beta_1 + n_{j1} - 1) +1, ... , \tau_\pi(\alpha_j \beta_j + \kappa + n_{jj} - 1) +1, ..., \tau_\pi(\alpha_L \beta_L + n_{jL} - 1) + 1))\\
\end{split}
\end{equation}

{\bf Inference of $\tau_\beta$ and $\tau_\pi$} 

To the best of our knowledge, there are no conjugate priors over a scaling factor for the parameters of a Dirichlet distribution, in the presence of an intercept. Hence, we estimate the next batch's learning rate using a Metropolis-Hastings (MH) jump. This approach is used in several other studies (such as \cite{knowles2011nonparametric} \cite{UsAISTATS14}) and is a valid MCMC move. 
For the MH step, one can choose a suitable candidate function and the samples are accepted with probability of acceptance $p(\xi \rightarrow \xi^*) \propto \min \left( 1, \frac{p(\xi^*) Q(\xi \rightarrow \xi^*)}{p(\xi) Q(\xi^* \rightarrow \xi)} \right)$.

To sample $\tau_\beta$, we have selected the candidate function as $G( \tau_\beta | \alpha, \beta)$, the prior over the learning rate. The new sample ($\tau^*_\beta$) is accepted with the probability in Equation~\ref{eq:MH}, updating $\tau_\beta$ for the current batch with the accepted sample. An identical approach can be taken for $\tau_\pi$ by replacing $\pi_j$ for $\beta$ in Equation~\ref{eq:MH}. The $\beta$ subscripts in $\tau_\beta$ are removed to avoid notational clutter.

\begin{equation}
\begin{split}
& p(\tau \rightarrow \tau^*) \propto \min \left( 1, \frac{p(\tau^*| \alpha, \beta) Q(\tau \rightarrow \tau^*)}{p(\tau| \alpha, \beta) Q(\tau^* \rightarrow \tau)} \right)\\
&\frac{p(\tau^*| \alpha, \beta) Q(\tau \rightarrow \tau^*)}{p(\tau| \alpha, \beta) Q(\tau^* \rightarrow \tau)} \propto \frac{ Dir(\beta | \alpha, \tau^*) G(\tau^*) G(\tau)}{Dir(\beta | \alpha, \tau) G(\tau) G(\tau^*)} =  \frac{ Dir(\beta | \alpha, \tau^*)}{Dir(\beta | \alpha, \tau)} \\
\end{split}
\label{eq:MH}
\end{equation}


\subsection{Discussion on the learning rates}

As per the above sections, the learning rates for each parameter are inferred separately to allow more degrees of freedom for independent adaptation of each parameter. The empirical results support this, as each of the learning rates ($\tau_A, \tau_\Sigma, \tau_\beta, \tau_\pi$) can adapt differently for the same sequence of data, depending on the complexity of the data and degree of evolution in the emissions and state transitions. Nevertheless, their impact pattern on the mean and covariance of the respective posterior distributions tends to be similar. As clearly shown for $\tau_\mu$ (Equation \ref{eq:tau_mu}) the learning rate does not change the mean, but reversely impacts the covariance (see Appendix D for more details). Accordingly, for all cases when $0 \leq \tau < 1$ the posterior distribution is more driven by the current observations. However, for $\tau > 1$ the inferred parameters follow the prior distribution more closely. In the following experiments, the dynamics of $\tau$ with respect to the data is explored more extensively.



\section{Experiments}
\label{sec:Experiments}

The experiments aim to explore the effectiveness of the proposed {\sc AdOn HDP-HMM} for segmentation and classification in a variety of scenarios. To closely examine the adaptability of the model, we have designed several synthetic datasets with stationary and evolutionary distributions. It also allows us to investigate the effects of using learning rates in enhancing adaptability, where an adaptive learning rate is noted concisely as `ada $\tau$` and the basic alternative with fixed learning rate is shown as $\tau = 1$. Following with two more video datasets, we demonstrate the performance of the proposed model in various challenging sequences with noisy data, abrupt changes and new classes in the test data. 
It is important to mention that the degree of challenge in the synthetic experiments is not easily comparable to the video data, due to differences in the nature of the signals, noise and, most importantly, degree of evolution that is stronger by design in the synthetic data. Hence, analysing both categories of experiments can shed more light on the adaptability of {\sc AdOn HDP-DHMM} in various contexts.

To evaluate the results more comprehensively, metrics for both classification and time segmentation performance are introduced. For classification accuracy, we have used frame-level comparison of the decoded classes with the ground truth (based on Hamming distance). To evaluate time segmentation, the standard metrics of precision and recall are utilised to indicate the accuracy of detecting boundaries between segments. A true boundary is regarded as correctly detected if a change of state is decoded within an interval of $\pm \Delta t$ frames from the ground truth location, where $\Delta t$ is set to 10 percent of the average segment length. Any additional detected boundaries are counted as false positives. We also report the difference between the overall number of actions detected in the test sequence and the number of actions in the ground truth (noted as \textit{cardinality}, with an ideal value of zero).

The empirical results are quantitatively reported in tables, also visualised in colour plots of ground-truth vs. estimated labels. In each illustration (for instance, Figure \ref{fig:evolutionary}), the horizontal axis is the time and the estimated labels are plotted on top of the true labels, providing a qualitative measure for the segmentation and classification performance. These plots are best viewed in colour.

\subsection{Synthetic data}
The basic framework of the synthetic dataset is generated from a univariate HMM, with 5 states distributed around dispersed means ($\mu = [ 100, 200, 300, 400, 500]$) with unit variance and a Dirichlet-distributed transition matrix ($\alpha = [3,3,3,3,3]$). This generative model is similar to the {\sc AdOn HDP-HMM}, but not an exact replicate, due to the absence of the HDP prior and adaptation of $\tau$ in the generative process (please refer to Figures \ref{fig:HDPHMM} and \ref{fig:onlineGM} for comparison).

\subsubsection{Stationary distributions}
Given the above basic configuration, the stationary experiments are run over 3 sequences of length 100, trained using leave-one-out cross validation. Hence, the distributions of training and test samples are the same. The test sequence is split into batches with approximate size of 16 time units. To provide adaptation, the inferred parameters of each batch are propagated into the next batch as priors.

\begin{figure}
\centering
\subfigure[]{\includegraphics[scale=0.45]{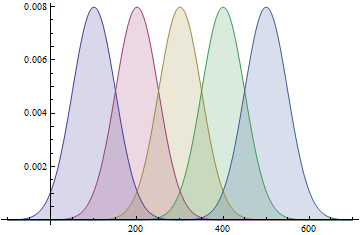}}
\subfigure[]{\includegraphics[scale=0.45]{SumofNoisyMixture.png}}
\caption{(a) The distribution of states in the noisy synthetic data set. Note that due to the large standard deviation ($\sigma = 50$) there is a significant overlap between the states, making recognition a challenging task.(b) The sum of density functions for all states.} \label{fig:noisyoverlap}
\end{figure}

The proposed Adaptive Online HDP-HMM is able to recognise and segment this basic version with 100 percent accuracy, whether or not the learning rates are used. To probe the model further, we add a significant noise to the above model by increasing the standard deviation to 50, thereby causing a considerable overlap between the distributions of each state (Figure \ref{fig:noisyoverlap}). Despite this substantial noise, the model is significantly accurate with an average of 76.3 percent frame-level accuracy. Repeating this experiment on the same data yet {\it with fixed learning rates} ($\tau = 1$), shows a noticeable decline in accuracy of 3 percentage points and undesirable extra states.
Table \ref{tab:Synthetic}, first two rows, shows the detailed accuracy figures in terms of precision, recall and number of inferred states.

\begin{table*}
\scriptsize
\begin{minipage}{\textwidth}
\begin{center}
\begin{tabular}{|l|l|l|l|l|}
\hline 
  & {\bf Accuracy} & {\bf Recall } & {\bf Precision} & {\bf Cardinality}  \T \\ [0.5ex]
\hline\hline
Stationary, Noisy ({\it ada} $\tau$) & {\bf  0.76}  & {\bf  0.92} &  0.92 & {\bf 0.33}  \T  \\
\hline 
Stationary, Noisy ($\tau = 1$) & 0.73      & 0.89        & {\bf 0.93}       &  1.7   \T \\
\hline\hline
Evolutionary, shifting mean ({\it ada} $\tau$) &  {\bf 0.97} & { 0.97} &  {0.99} &  {\bf 0}  \T  \\
\hline 
Evolutionary, shifting mean ($\tau = 1$) &  {0.71} &  {{\bf 0.99}} &  {{\bf 1}} &  {1}   \T \\
\hline\hline
Evolutionary, new class ({\it ada} $\tau$) &  {\bf 1.00} & {{\bf 1.00}} &  {{\bf 1.00}} &  {\bf 0}  \T  \\
\hline 
Evolutionary, new class ($\tau = 1$) &  {0.86} &  {0.86} &  {0.98} &  {0} \T \\
\hline\hline
Evolutionary, combined ({\it ada} $\tau$) &  {\bf 0.93} & {{\bf 1.00}} &  {{\bf 1.00}} &  {\bf 1}   \T  \\
\hline 
Evolutionary, combined ($\tau = 1$) &  {0.81} &  {0.95} &  {0.97} &  {2}   \T \\
\hline
\end{tabular}
\end{center}
\caption{Frame-level accuracy, segmentation recall, precision and state cardinality error for the synthetic experiments. Each table section includes the respective results of experiment secions, comparing the performance with and without the learning rate: i) Stationary with high noise reporting average results on 3 sequences, ii) Evolutionary with shifting means, iii) Evolutionary with new states, iv) Evolutionary with combined shifts and new states.}
\label {tab:Synthetic}
\end{minipage}
\end{table*}

\begin{figure}
\centering
\subfigure[shifting mean ({\it ada} $\tau$)]{\includegraphics[scale=0.25]{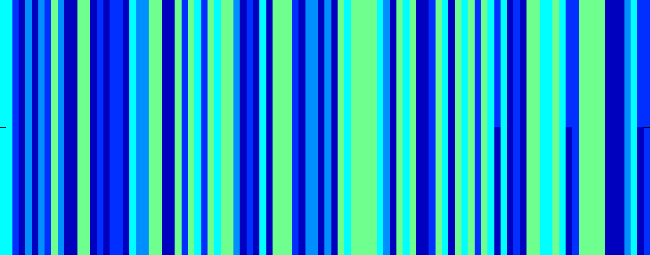}} \hspace{0.25cm}
\subfigure[new class ($\tau = 1$)]{\includegraphics[scale=0.25]{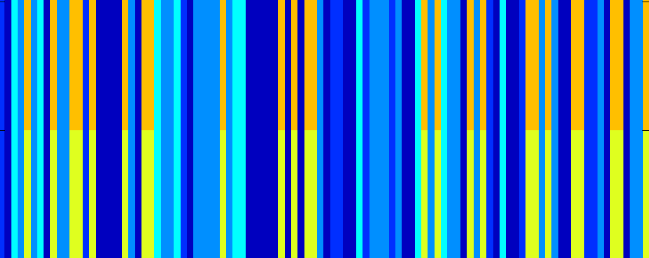}} \hspace{0.25cm}
\subfigure[combined ({\it ada} $\tau$)]{\includegraphics[scale=0.25]{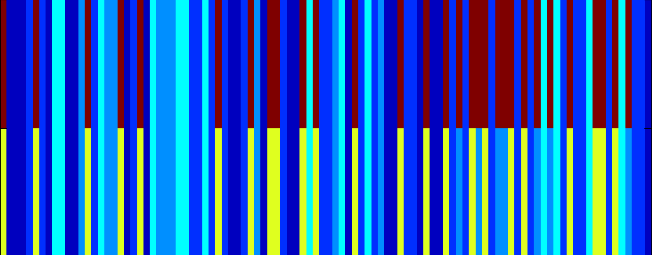}}
\subfigure[shifting mean ($\tau = 1$)]{\includegraphics[scale=0.25]{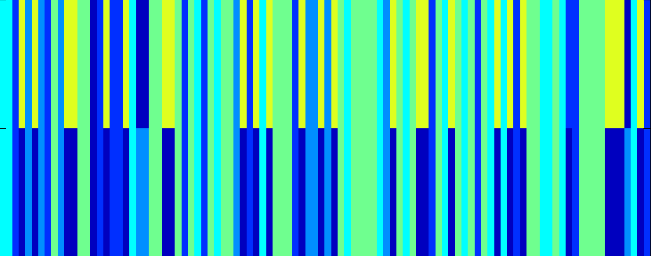}} \hspace{0.25cm}
\subfigure[new class ({\it ada} $\tau$)]{\includegraphics[scale=0.25]{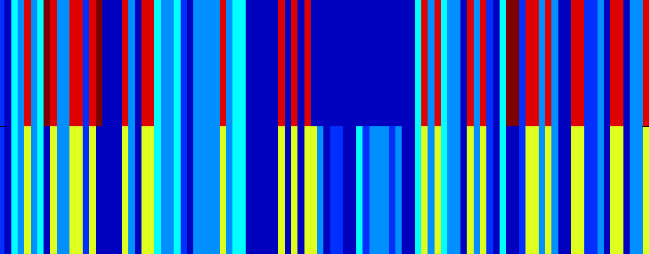}} \hspace{0.25cm}
\subfigure[combined ($\tau = 1$)]{\includegraphics[scale=0.25]{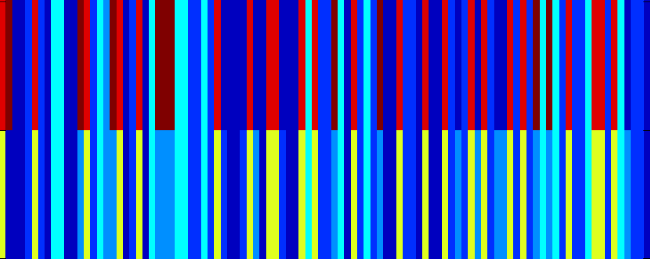}}
\caption{Segmentation and classification results for evolutionary synthetic data, using fixed ($\tau = 1$) and adaptive ({\it ada} $\tau$) learning rates. Top half of the stripes: \textit{predicted states}; bottom half: \textit{ground truth}. (a,d) Shifting means: Without the adapting effects of the learning rate, shifting means can be misclassified as new classes (yellow) in d. (b,e) New class: is shown in yellow in the ground truth. The new class has been recognised and learnt in both cases, assigned a random colour (orange in (b) and red in (e)). (c,f) Combined: Adding both challenges causes slight decrease in accuracy and an extra new class. Nevertheless, the considerable performance gap in utilising $\tau$ is still visible.} 
\label{fig:evolutionary}
\end{figure}

\subsubsection{Evolutionary distributions}
A more advanced experiment is designed by training the model on synthetic data with evolving distributions, either involving gradual shifts to the means of each class or including new unseen classes. The standard deviation for this experiment is set to $\sigma = 10$.

{\bf Shifting class means: }
To examine the adaptability of the model, we drift the class means by $\delta = 0.5$ at each time step. Therefore, an instance appearing at $t = 10$ in the test sequence is generated from a distribution with its mean shifted by 5 units. For a non-adaptive model and given the synthetic generation scheme, such data can cause significant classification errors after a few tens of time units. However, the results of the Adaptive Online HDP-HMM demonstrate smooth adaptation and excellent accuracy over the evolving sequence (Figure \ref{fig:evolutionary}). There are a few misclassifications towards the end of the sequence which are due to the heavy distributional drift. Comparison between these results and that of {\it fixed} learning rates shows a significant drop of 26 percentage points in accuracy and one undesirable new class (see Table \ref{tab:Synthetic}, section ii).

{\bf New classes: }
In this experiment, distributions do not shift, yet one new class appears around $\mu = 600$ with the same $\sigma$ as the other classes. The model is able to create a new state (shown with a random new colour in Figure \ref{fig:evolutionary}), learn and consistently recognise it in the later batches without distorting parameters of the existing classes. The overall accuracy of 100 percent for this experiment is mostly thanks to the contribution of the learning rate in adjusting the variances of each class with respect to the degree of adaptation. Not using learning rates can highly reduce accuracy (14 percentage points) due to drift in the existing classes (Table \ref{tab:Synthetic}), also exhibiting one extra class and  reduced recall and precision.

{\bf Combination of the two: }
Combining the above two evolutionary scenarios, we test the proposed model on a sequence with a {\it new} class that needs to be distinguished among the existing {\it shifting} classes. The challenge is two-fold: i) the shifting modes are prone to being misclassified as new classes, and ii) the new class might be merged into one of the existing shifted modes. This experiment is the closest to challenging real world scenarios where new states are likely to appear while the distributions can change over time. Given the combined challenge, the {\sc AdOn HDP-HMM} proves highly accurate (93 percent), exhibiting a considerable improvement on the accuracy (12 percentage points) and cardinality of states thanks to the learning rate mechanism. 

The performance of the Adaptive Online HDP-HMM is not perturbed by these challenges because the learning rate tunes the adaptability of the parameters with respect to the observed data. In an evolutionary scenario, the likelihood of the observations given the current parameters is low. This causes the posterior covariance learning rate ($\tau_{\Sigma}$) to increase, keeping the variance close to its prior. This, in turn, prevents a drift of the variance towards large values and allows for the mean to evolve. The concentration of $\tau_{\mu}$ around zero is an empirical support for this claim (see Figure \ref{fig:samplerPerf}b). 

In the absence of the learning rate, the model still learns and recognises the new state thanks to properties of HDP. However, the overall performance deteriorates. On the one hand,  new undesirable classes appear in response to drift. On the other, some of the existing classes collapse into a single one, due to considerable increase of variance caused by the class shifts. This rigid increase in variance does not allow the means to evolve, ultimately forcing the model to merge some of the neighboring states into a single class with a large variance (Figure \ref{fig:evolutionary}e,f).


\subsection{ Activity recognition datasets}
In this section, we use two video datasets to assess the performance of the proposed model in activity recognition scenarios.

\subsubsection{ Collated Weizmann dataset}
The Weizmann dataset contains 93 single-action videos from a set of 10 classes performed by 9 different actors. While the recognition accuracy on the original dataset is saturated \cite{WangSuter2007} \cite{Nanni2011}, some studies have collated its individual actions into (unsegmented) sequences to experiment with time segmentation \cite{HoaiLD11}. In a similar way, we have created 4 sequences, each consisting of 12 random actions selected from the provided action classes. Each sequence consists of approximately 900 frames. 
As feature set, we have used the position of the actor's centroid in the image plane and the distances between the centroid and the actors' contour along five given directions~\cite{Moghaddam2009}.

\begin{table*}
\scriptsize
\begin{minipage}{\textwidth}
\begin{center}
\begin{tabular}{|l|l|l|l|l|l|l|l|l|l|l|l|l|}
\hline 
  & \multicolumn{4}{|c|}{\bf Accuracy} &  \multicolumn{4}{|c|}{\bf F1 score} & \multicolumn{4}{|c|}{\bf Cardinality} \T \\ [0.5ex]
\hline\hline
 {\bf Method} & {\bf S1}& {\bf S2} & {\bf S3} & {\bf S4} &  {\bf S1}& {\bf S2} & {\bf S3} & {\bf S4}& {\bf S1}& {\bf S2} & {\bf S3} & {\bf S4} \T  \\
\hline
{\it Online} HDP-HMM ({\it ada} $\tau$) & {\bf 0.82} & {\bf  0.76} & 0.89 & {\bf 0.81} &  0.92 & 0.66 & 0.95 & 0.80 & {\bf 0} & {\bf 0} & {\bf 0} & {\bf 0} \T  \\
\hline 
{\it Online} HDP-HMM ($\tau = 1$) & 0.81 & 0.70 & {\bf 0.95} & 0.80  &  0.92 & 0.66  & 0.89 & 0.80    & 0 & -1 & 0 & -1 \T \\
\hline
{\it Offline} HDP-HMM & {\it 0.78} & {\it 0.76} & {\it 0.95} & {\it 0.81} & {\it 0.91} &{\it 0.66} &{\it 0.95} &{\it 0.81 } &{\it0}&{\it0} &{\it0 }&{\it0} \T  \\
\hline
{\it Offline} Max-margin \cite{HoaiLD11} & \multicolumn{4}{|c|} {\it 0.87 (avg)} &  \multicolumn{4}{|c|} {-} & \multicolumn{4}{|c|} {-} \T \\
\hline
\end{tabular}
\end{center}
\caption{Frame-level accuracy, segmentation F1 score, and difference in decoded state cardinality for the adaptive online HDP-HMM variants and state-of-the-art studies on the collated Weizmann dataset.}
\label {tab:Weizmann}
\end{minipage}
\end{table*}

\begin{figure*}
\begin{center}
\subfigure[Sequence 1]{\includegraphics[scale=0.3, clip]{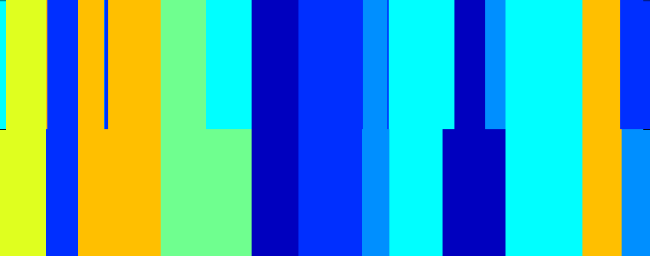}} \hspace{1mm}
\subfigure[Sequence 2]{\includegraphics[scale=0.3, clip]{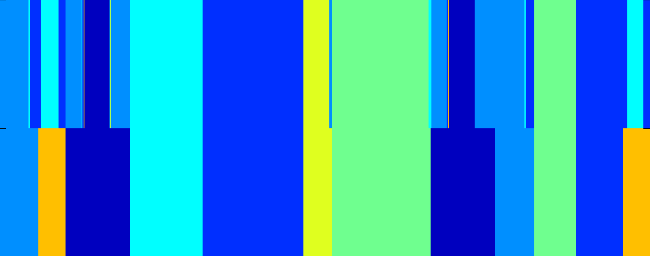}} \hspace{1mm}
\subfigure[Sequence 3]{\includegraphics[scale=0.3, clip]{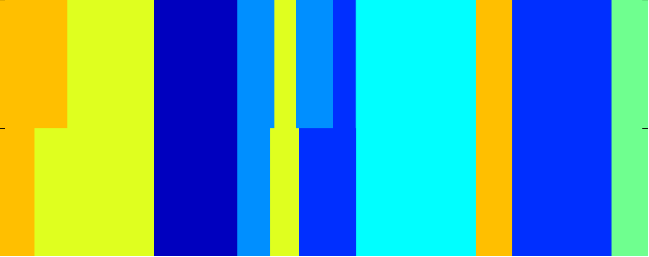}}\hspace{1mm}
\subfigure[Sequence 4]{\includegraphics[scale=0.3, clip]{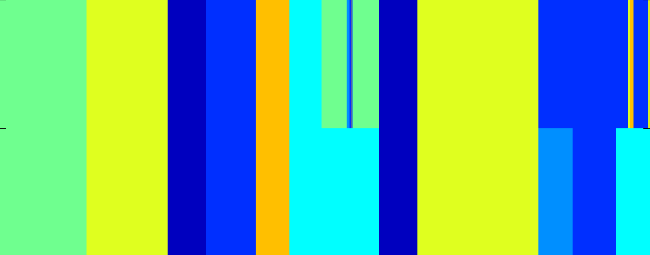}} \hspace{1mm}
\end{center}
   \caption{Estimated states for Weizmann dataset. Action labels are represented by colours.}
\label{fig:Weizmann}
\end{figure*}

The estimated states of the {\sc AdOn HDP-HMM} variants over the above sequences are visualised in Figure~\ref{fig:Weizmann}, showing remarkable qualitative accuracy in segmentation and classification. The quantitative results are reported in Table \ref{tab:Weizmann}, including an {\it Offline} variant representing the experiment with a single batch including the whole test sequence. This variant is run for the sake of comparison with a similar offline max-margin study \cite{HoaiLD11}. However, the results are not directly comparable for two reasons: a) the datasets are similar in conception, yet different in sequence collation, and b) the classifier in \cite{HoaiLD11} operates over a closed set of classes, as opposed to ours that allows unlimited number of classes. 
The results with the fixed learning rate ($\tau = 1$) show a similar trend to the adaptive, and only a slightly lower average accuracy. This can be due to the stationary nature of the dataset, as training and test sequences are drawn from similar distributions and adaptation is not significant. In addition, the accuracy with the online processing does not show any noticeable deterioration over the full, offline processing.

\subsubsection{ TUM kitchen dataset}
The TUM kitchen dataset is a human assistive dataset, consisting of natural unsegmented sequences of everyday activities performed in a typical kitchen environment \cite{tenorth09dataset}. The dataset contains multi-modal data, annotated separately for the actors' left and right hands (9 classes) and torso (2 classes). The features are 28D vectors of joint coordinates for the torso and the relevant hands. The main actions include `Reaching', `Releasing Grasp Of Something', `Taking An Object', `Reaching Upward', `Lowering An Object', opening and closing doors and drawers and `Carrying While Locomoting', the distinction of which are quite subtle at times even for human annotators. The main advantage of this dataset over the collated Weizmann is that the transitions between actions occur naturally and the boundaries are vague even to human annotation, hence time segmentation is more challenging.

In our experiments, we have performed segmentation and classification on the actions of the left and right hands, separately. All the sequences provided by the 3D motion capture sensors are used in leave-one-out cross validation tests. Experiments are run for both the typical sequences (denoted as `{\it robotic}', taking objects one by one), and the more challenging ones (`{\it complex}' including sequences with multiple objects moved together, in arbitrary order and repeatedly).

For a general study of performance, we run an experiment on all the above sequences involving both the {\it robotic} and {\it complex}. The difference between them is in state transition probabilities, height and size of actors and frequencies of action occurrence. The experiment is repeated with fixed and adaptive learning rates and results are compared in Table \ref{tab:TUMresults}, generally showing significant match in frame-level accuracy. The comparison between similar sequences with fixed vs. adaptive learning rate shows a minor improvement of frame-level accuracy and significant decrease of state cardinality error. Note that the figures under cardinality show differences between inferred vs. actual number of states. To facilitate visual evaluation, 4 of the sequences are colour-plotted in Figure \ref{fig:TUMplot}. It is worth noting that classes in this dataset may prove hard to segment. For instance, distinction between putting object on the table and leaving grasp of it can be very subtle (the back-to-back lavender-blue and light-blue colours in Figure \ref{fig:TUMplot}). This becomes more challenging when a model has extra degrees of freedom for deriving a dynamic number of classes and explains the negative cardinality in the results.

To specifically observe the adaptive behaviour, we have trained the model on the {\it robotic} sequences and tested it on {\it complex} ones. Although the emission parameters might not radically change in this scenario, the transition probabilities need to adapt due to changes in the order of actions in the {\it complex} set. Table \ref{tab:TUMcomplex} can be used to observe the remarkable contribution of the learning rate mainly in cardinality and overall accuracy. Similar to the synthetic results, in the presence of learning rates the model is able to prevent an excessive increase of the variance and avoid neighboring classes to collapse into one (the phenomena that can be observed when $\tau = 1$  in Figures~\ref{fig:TUMplot}d,e).

To evaluate the ability to recognise new classes, we have taken the first 4 sequences and removed the observations related to `Lowering an object' (shown in lavender-blue in Figure \ref{fig:TUMplot}f) in all but the first sequence. We have then trained the model on sequences 2-4 and tested on the sequence containing the new action. {\sc AdOn HDP-HMM} is able to recognise a new action ({brown in Figure \ref{fig:TUMplot}f) and learn its parameters with consistent future recognition. This significant property of the model is inherent to the HDP approach and the behaviour is similar, irrespective of whether or not the learning rate is utilised.

The closest study on the TUM kitchen dataset leverages a CRF \cite{tenorth09dataset}. This method is not directly comparable to ours since {\sc AdOn HDP-HMM} is online, adaptive and with a dynamic class set. To create a closer match, we have run the {\it Offline} variant of {\sc AdOn HDP-HMM}, the results of which are similar to the CRF and outperforming it for complex sequences. This finding aligns with our principal claim that adaptability leads to remarkable improvements when the test distributions are different from the training. The distribution of $\tau_\pi$ and $\tau_\beta$ (the transition-related learning rates) for these experiments are mainly peaked around 0.1, indicating that the learning rates encourage the model to rely on the observed data to infer the HDP transition probabilities, which translates into more adaptability.

\begin{table}
\begin{center}
\begin{tabular}{|l|l|l|l|l|l|l|l|l|l|}
\hline 
  & \multicolumn{4}{|c|}{\bf Accuracy} & \multicolumn{4}{|c|}{\bf Cardinality} \T\\ [0.5ex]
\hline \hline
 \multirow{2}{*}{\bf Sequences}  & \multicolumn{2}{|c|}{\bf RH} &  \multicolumn{2}{|c|}{\bf LH}  & \multicolumn{2}{|c|}{\bf RH} &  \multicolumn{2}{|c|}{\bf LH} \T \\  \cline{2-9}
   & {\it ada $\tau$}& { $\tau=1$} & {\it ada $\tau$}& { $\tau=1$} & {\it ada $\tau$}& { $\tau=1$} & {\it ada $\tau$}& { $\tau=1$}\\
\hline
\textit{Online} Seq 0-0  & 0.79 & 0.81  & 0.73 & 0.71 & 0 & -1  & -1 & -1  \T \\ 
\textit{Online} Seq 0-1  &  0.79 & 0.82&  0.75 & 0.75 & -2 & -1 & 0 & -1  \T \\ 
\textit{Online} Seq 0-2  &  0.76 & 0.70 & 0.78 & 0.75 & -2 & -1 &-1 & -1  \T \\ 
\textit{Online} Seq 0-3  &  0.84 & 0.84 & 0.67 & 0.69 & 0 & -2 &0 & -1  \T \\ 
\textit{Online} Seq 0-4  &  0.70 & 0.69 &0.71 & 0.72 & 1 & -1 & -2 & -3  \T \\ 
\textit{Online} Seq 0-6  &  0.51 & 0.48 & 0.56 & 0.55 & -3 & -6 & -1 & -3  \T \\ 
\textit{Online} Seq 0-7  &  0.45 & 0.48 & 0.57 & 0.55 & -3 & -4 & -1 & -3  \T \\ 
\textit{Online} Seq 0-8  &  0.64 & 0.68 & 0.62 & 0.63 & -1 & -3 & -2 & -2  \T \\ 
\textit{Online} Seq 0-9  &  0.73 & 0.71 &0.70 & 0.69 & 0 & -2 &-1 & -2  \T \\ 
\textit{Online} Seq 0-10  &  0.79 & 0.79 & 0.68 & 0.70 & 0 & -2 &0 & -1  \T \\ 
\textit{Online} Seq 0-11  &  0.70 & 0.76 & 0.63 & 0.63 & -5 & -4 & -2 & -3  \T \\ 
\textit{Online} Seq 0-12  &  0.64 & 0.64 & 0.58 & 0.55 & -1 & -2 & -3 & -5 \T \\ 
\textit{Online} Seq 1-0  &  0.65 & 0.69 &0.68 & 0.69 & -1 & -2 &-3 & -4  \T \\ 
\textit{Online} Seq 1-1  &  0.71 & 0.69 & 0.65 & 0.62 & -1 & -1 & -2 & -3  \T \\ 
\textit{Online} Seq 1-2  &  0.63 & 0.63 &0.76 & 0.74 & -1 & -1 & 0 & -1  \T \\ 
\textit{Online} Seq 1-3  &  0.14 & 0.14 & 0.66 & 0.65 & -6 & -6 & 0 & 0  \T \\ 
\textit{Online} Seq 1-4  &  0.64 & 0.67 & 0.74 & 0.71 & -4 & -5 & 0 & -1  \T \\ 
\textit{Online} Seq 1-5  &  0.67 & 0.67 & 0.61 & 0.61 & 0 & 0 & -2 & -2  \T \\ 
\textit{Online} Seq 1-7  &  0.69 & 0.68 & 0.60 & 0.58 & -1 & -1 & -1 & -2  \T \\ 
\hline \hline
\multicolumn{9}{|c|}{\it robotic sequences} \\
\hline
Avg \textit{Online}  &  {\bf 0.80}  & 0.79  & {\bf 0.73} &   {\bf 0.73} & {\bf 1.00}  & 1.25  & {\bf 0.5} & 1.00  \T \\
\hline
Avg \textit{Offline}  &  0.80  & {\bf 0.81}  & {\bf 0.74} &   0.73 & {\bf 1.00}  & 1.50  & {\bf 1.00} & 1.10  \T \\
\hline
Avg \textit{Offline} CRF ~\cite{tenorth09dataset} & \multicolumn{4}{|c|}{0.83  ({\it avg})}  & \multicolumn{4}{|c|}{-} \T\\
\hline \hline
\multicolumn{9}{|c|}{\it complex sequences} \\
\hline
Avg \textit{Online}  &  {\bf 0.66}  & {\bf 0.66}  & {\bf 0.67} &  0.66 & {\bf 1.68}  & 2.37  & {\bf 1.16} & 2.26 \T \\
\hline
Avg \textit{Offline}  &  {\bf 0.66}  & {\bf 0.66}  & {\bf 0.67} &  0.66 & {\bf 1.48}  & 2.28  & {\bf 1.23} & 2.26 \T \\
\hline
Avg \textit{Offline} CRF ~\cite{tenorth09dataset} & \multicolumn{4}{|c|}{0.63 ({\it avg})}  & \multicolumn{4}{|c|}{-} \T\\
\hline
\end{tabular}
\end{center}
\vspace{1mm}
\caption{Frame-level accuracy and state cardinality error for Adaptive Online HDP-HMM on all TUM kitchen sequences. The comparison between similar sequences with fixed and adaptive learning rate ($\tau$) shows incremental improvement on frame-level accuracy and significant decrease on state cardinality error. Note that the figures under cardinality show differences between inferred vs. actual number of states, considering the sign. However the absolute values are utilised to calculate the average cardinality error.}
\label {tab:TUMresults}
\end{table}

\begin{figure*}
\begin{center}
\subfigure[Seq 0-3, RH, {\it ada} $\tau$]{\includegraphics[scale=0.25, clip]{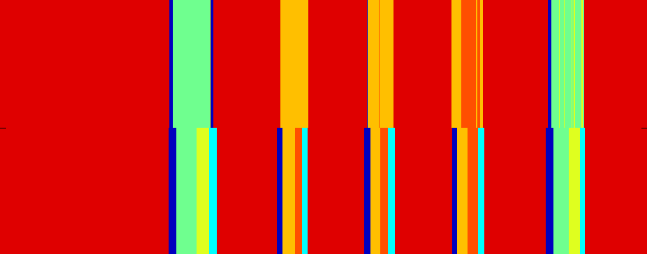}} \hspace{1mm}
\subfigure[Seq 0-4, LH, {\it ada} $\tau$]{\includegraphics[scale=0.25, clip]{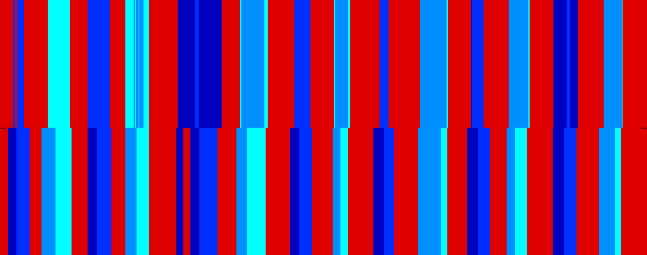}} \hspace{1mm}
\subfigure[Seq 1-3, LH, {\it ada} $\tau$]{\includegraphics[scale=0.25, clip]{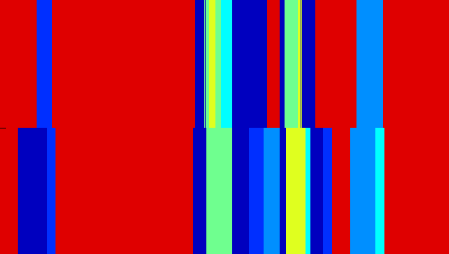}}\hspace{1mm}
\subfigure[Seq 0-3, RH, $\tau = 1$]{\includegraphics[scale=0.25, clip]{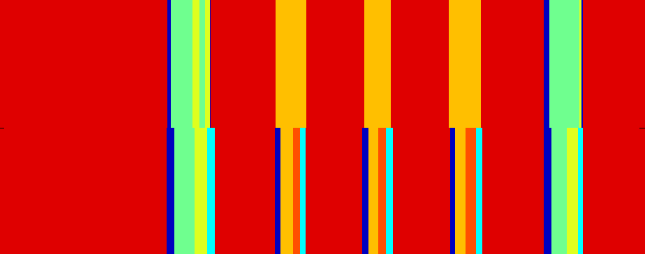}} \hspace{1mm}
\subfigure[Seq 0-4, LH, $\tau = 1$]{\includegraphics[scale=0.25, clip]{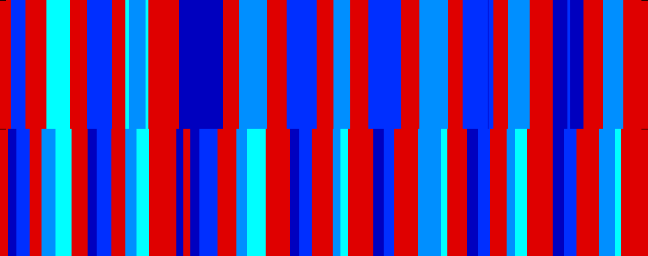}} \hspace{1mm}
\subfigure[Seq 0-0, {\bf new class}]{\includegraphics[scale=0.25, clip]{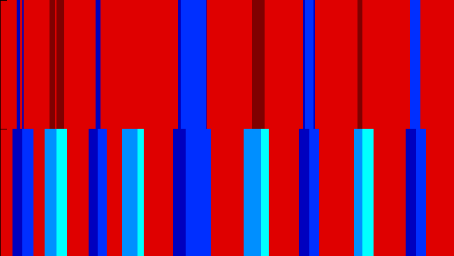}}\hspace{1mm}
\end{center}
\caption{Estimated states for the TUM kitchen dataset using the Adaptive Online HDP-HMM. LH and RH stand for left and right hand. The first two columns show {\it robotic} sequences, whereas the third column includes {\it complex} ones. (c) is a {\it human} sequence with altered orders of actions performed spontaneously and (d) contains a new action shown in lavender-blue in the ground-truth and recognised by the model in a random brown colour. In most cases, using adaptive $\tau$ causes noticeable improvements on both the performance and cardinality of inferred states. }
\label{fig:TUMplot}
\end{figure*}

\begin{table}
\begin{center}
\begin{tabular}{|l|l|l|l|l|l|l|l|l|l|}
\hline 
  & \multicolumn{4}{|c|}{\bf Accuracy} & \multicolumn{4}{|c|}{\bf Cardinality} \T\\ [0.5ex]
\hline \hline
 \multirow{2}{*}{\bf Sequences}  & \multicolumn{2}{|c|}{\bf RH} &  \multicolumn{2}{|c|}{\bf LH}  & \multicolumn{2}{|c|}{\bf RH} &  \multicolumn{2}{|c|}{\bf LH} \T \\  \cline{2-9}
   & {\it ada $\tau$}& { $\tau = 1$} & {\it ada $\tau$}& {$\tau = 1$} & {\it ada $\tau$}& {$\tau = 1$} & {\it ada $\tau$}& {$\tau = 1$}\\
\hline
\textit{Online} Actor1, complex  & {\bf 0.73} & 0.72  & 0.65 & {\bf 0.68} & {\bf -2} & -3  & {\bf 2} & {\bf -2}  \T \\ 
\textit{Online} Actor3, complex  &  {\bf 0.55} & 0.54 &  {\bf 0.52} & 0.49 & {\bf -1} & -3 & {\bf -4} & -6  \T \\ 
\textit{Online} Actor1, repetitive  &   0.45 & {\bf 0.48} & {\bf 0.57} & 0.55 & {\bf -3} & -4 & {\bf -1} & -3  \T \\
\hline
\end{tabular}
\end{center}
\vspace{1mm}
\caption{Adaptability experiment: frame-level accuracy and state cardinality error for Adaptive Online HDP-HMM trained with the {\it robotic} sequences and tested on the {\it complex} ones. The comparison between similar sequences with/without learning rate ($\tau$) shows noticeable improvement on frame-level accuracy and significant decrease on state cardinality error.}
\label {tab:TUMcomplex}
\end{table}


\subsection{ Sampling efficiency and computational time}

We next examine the Gibbs sampler's mixing rate and execution time for the above experiments. To gain an overall understanding of parameter mixing (emission and transition) the log-likelihood is shown in Figure \ref{fig:samplerPerf}e. Since most of the sampled variables contribute to the likelihood calculation, the well-mixed results indicate general mixing efficiency in the model. Additionally, mixing trends of the learning rates  ($\tau_\mu, \tau_\Sigma, \tau_\pi,\tau_\beta$) for a generic evolutionary run are shown Figure \ref{fig:samplerPerf}a-d, both to monitor mixing and support the experiments' discussion. The large values of $\tau_\Sigma$ prevents the model from the immediate tendency to increase the variance to fit the changing distributions. Rather, the model allows for the means to evolve, by converging to small values of $\tau_\mu$. The similarly small values of $\tau_\pi$ and $\tau_\beta$ ensure adaptability of the model towards changing state transitions for HDP-HMM. Through the orchestration of these parameters, the proposed model can adapt to changes in the streaming batches with more exact account of the true cardinality of the classes and be immune from collapsing neighboring classes into a single one.

Eventually, the computational time per frame for runs on an Intel Xeon E5 2.90 GHz processor, over the Weizmann and TUM kitchen datasets are shown in Figure \ref{fig:samplerPerf}f. The boxplot includes online and offline variants, with and without learning rates to help explore how using the learning rates and online scheme can affect the computational time. Based on the elapsed time (in seconds), the offline run is the fastest since all the data are processed in a single batch. The adaptive online runs occur in 3-4 batches of 1000 iterations each, therefore indicate an increase of about 5-10 ms in completion time. Adapting the learning rate can cause between 3-10 ms delay, yet given the discussed benefits particularly for evolving sequences, this latency is quite reasonable. It is important to mention that given the initial bootstrap training, the Gibbs algorithm converges rapidly allowing for the model to run in acceptable time. Overall, using the learning rate ensures multiple improvements without imposing excessive computational load on the system.

\begin{figure*}
\begin{center}
\subfigure[$\tau_\mu$]{\includegraphics[scale=0.25, clip]{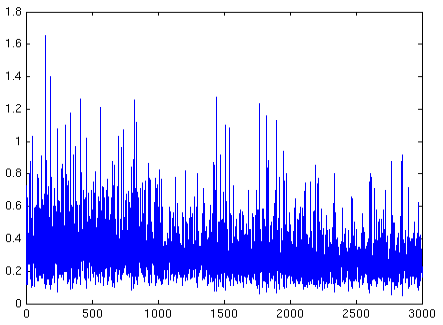}} 
\subfigure[$\tau_\Sigma$]{\includegraphics[scale=0.25, clip]{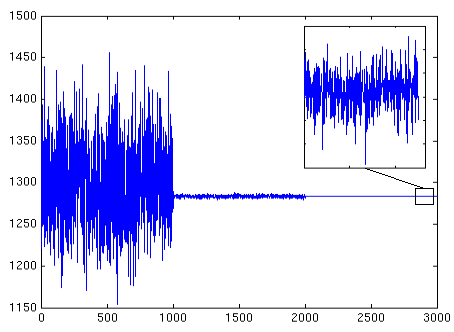}}
\subfigure[$\tau_\pi$]{\includegraphics[scale=0.25, clip]{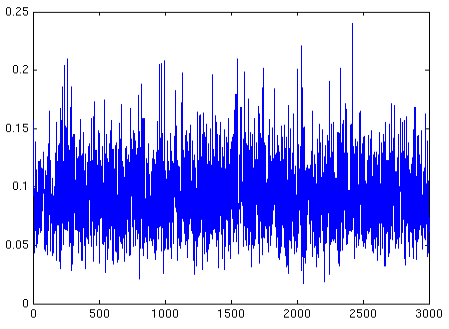}} 
\subfigure[$\tau_\beta$]{\includegraphics[scale=0.25, clip]{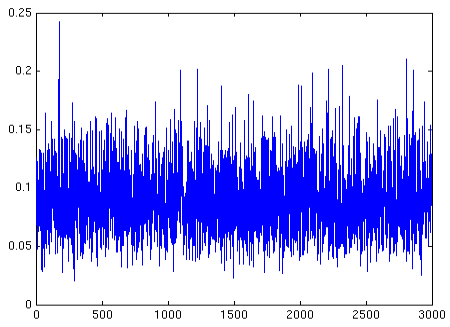}} 
\subfigure[Loglikelihood]{\includegraphics[scale=0.35, clip]{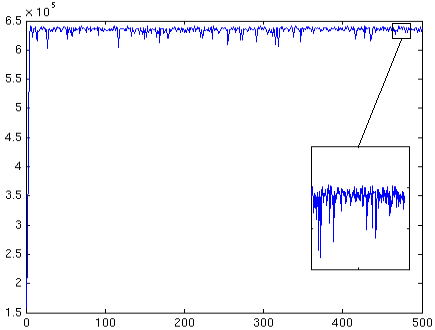}} 
\subfigure[Computational time]{\includegraphics[scale=0.35, clip]{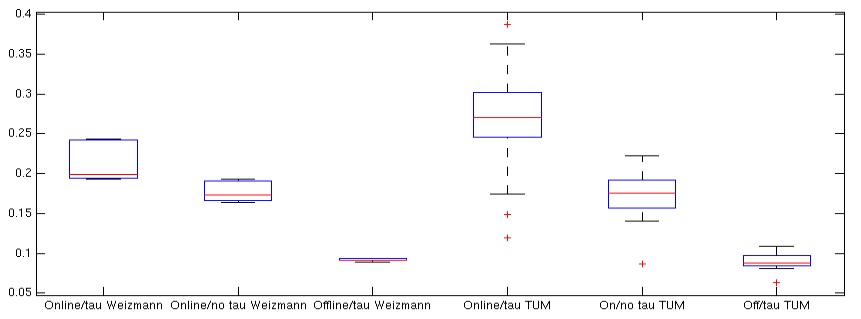}} 
\end{center}
\caption{Sampling efficiency and computational time: (a-d) Sample mixing for all the 4 proposed learning rates across three online batches and (e) Log likelihood plot for first batch of a Weizmann run, show well mixing and convergence both for the learning rates and generally for all other parameters involved in likelihood calculation. (f) Computation time per frame (seconds) for Weizmann and TUM kitchen datasets, over the online and offline variants, with/without learning rates.}
\label{fig:samplerPerf}
\end{figure*}


\section{Conclusion}
\label{sec:conclusion}

In this paper, we have proposed a novel, adaptive online model suited for on-the-fly time segmentation and recognition of sequential data. The proposed {\sc AdOn HDP-HMM} is capable of online segmentation and classification of streaming batches of data over incremental class sets. The main contribution of this model is the unsupervised posterior adaptation of the parameters over the successive data batches. This is accomplished by using a learning rate that dynamically tunes the model balancing the impact of the current batch with the memory accumulated so far. This proves an effective solution for online sequential estimation problems requiring adaptation over evolving distributions.

The performance of {\sc AdOn HDP-HMM} is evaluated via a number of experiments including {\it stationary} and {\it evolutionary} scenarios. Thereby, we have tested the general segmentation and classification accuracies in addition to the ability to detect the correct number of classes. The results are reported on variations of synthetic data and two activity recognition video datasets (Collated Weizmann and TUM Assistive Kitchen). The proposed model has achieved a remarkable accuracy in all cases, and considerable improvements in evolutionary scenarios.

Thanks to the unsupervised adaptive online estimation and the capacity to learn over infinite class sets, the proposed {\sc AdOn HDP-HMM} can be a solution for sequential estimation in a number of scenarios which have received relatively little attention in the literature. Not relying on human intervention, revision or correction of estimated labels, this model can be a suitable candidate for streaming applications. In addition, although designed for evolutionary distributions, its accuracy over stationary data has proved higher than or equal to that of the most comparable results, and the computational load is not affected significantly.


\section {Appendix A: The balancing effect of $\tau$}

In this section we address posterior inference of parameters and explore how the prior and likelihood distributions convey the knowledge of the current observations and accumulated summary of previous data. Considering the online HDP-HMM model with parameters $\phi$, observations $Y$ and learning rate $\tau$, the posterior for parameters in the $n^{th}$ batch is:

\begin{equation}
\begin{split}
& p(\phi_{n}|Y_{n}, \phi_{n-1}, \tau) \propto p(Y_{n}|\phi_{n-1}) p(\phi_{n-1}) ^\tau   \\
& \propto \underbrace{\prod^N_{i = 1}{p(y_{n,i} | \phi_{n-1})}}    \underbrace{\left( p(\phi_0)  \prod^{n-1}_{j=1}{ \prod^N_{i=1} {p(y_{j,i} | \phi_{j-1})}} \right)^\tau } \\
& \hspace{1.5cm} N \hspace{2.7cm} N(n-1)\tau
\end{split}
\end{equation}

As more batches stream in (i.e. $n$ increases), the weight of prior is accumulated and adaptivity to new data declines. The learning rate, however, can be used as an equaliser that controls the balance of prior versus likelihood and tunes model adaptivity. For positive values of $\tau < 1$, the model discounts the impact of accumulated previous data and allows for more adaptivity. However, when $\tau > 1$, posterior $\phi_n$ is inclined to follow the prior more strictly. In other words, $\tau$ can be seen as the scaling coefficient for the number of {\it `pseudo-observations'} in the prior.


\section {Appendix B: Conjugacy for $\tau_\Sigma$}

To sample $\tau_\Sigma$ from the posterior, ideally we would like to consider a conjugate prior that analytically derives the posterior hyper-parameters, given those of the prior and the sufficient statistics of the current data. A candidate prior for the IW distribution is the Gamma. In this section, we investigate if the Gamma can be proven a conjugate prior for the IW likelihood, considering the impact of the learning rates on $\Psi'$ and $\nu'$. 

Given the proposed learning rate model, the probability density function of the Inverse-Wishart distribution can be redefined as below. We have derived the new hyper-parameters through conversion to canonical parameters, multiplication with the learning rate and reversion to the standard form to simplify sampling.

\begin{equation}
\begin{split}
&\Psi' = \tau \Psi \hspace{2cm}  \nu' = \tau (\nu + p + 1) - p - 1 = c\tau + c' \\
&IW_p (\Psi, \nu, \tau) = \frac{|\Psi|^\frac{c\tau + c'}{2}}{2^{\frac{c\tau p + pc'}{2}} \Gamma_p(\frac{c\tau + c'}{2})}  |Y|^{-\frac{c\tau + c' + p + 1}{2}} \exp\left( {-\frac{1}{2}} tr(\tau \Psi Y^{-1}) \right)\\
\end{split}
\end{equation}

We assume Gamma is  the conjugate prior distribution for sampling $\tau$, and try to prove it below.

\begin{equation}
\begin{split}
&G(\tau | \Sigma,\alpha^*, \beta^*) \propto IW(\Sigma | \Psi, \nu, \tau) G(\tau | \alpha, \beta)\\
&G(\tau | \Sigma,\alpha^*, \beta^*) \propto \frac{|\Psi|^\frac{c\tau + c'}{2}}{2^{\frac{c\tau p + pc'}{2}} \Gamma_p(\frac{c\tau + c'}{2})}  |\Sigma|^{-\frac{c\tau + c' + p + 1}{2}} \exp\left( {-\frac{1}{2}} tr(\tau \Psi \Sigma^{-1}) \right) \frac{\beta ^ \alpha}{\Gamma(\alpha)} \tau ^ {\alpha - 1}\exp(-\beta \tau)\\
\end{split}
\end{equation}

Thanks to proportionality, we can remove the constant terms with respect to $\tau$:

\begin{equation}
\begin{split}
&G(\tau | \Sigma,\alpha^*, \beta^*) \propto \frac{|\Psi|^\frac{c\tau}{2}}{2^{\frac{c\tau p}{2}} \Gamma_p(\frac{c\tau}{2})}  |\Sigma|^{-\frac{c\tau}{2}} \exp\left( {-\frac{1}{2}} tr(\tau \Psi \Sigma^{-1}) \right) \tau ^ {\alpha - 1}\exp(-\beta \tau)\\
\end{split}
\end{equation}

Ideally, we should create terms proportional to `$ \tau^{\alpha-1} \exp\left( - \tau\ \left( \frac{1}{2} \text{tr}(\Psi^{-1} Y) + \beta \right)  \right)$', but because $\tau$ affects both hyper-parameters, the initial term related to the degrees of freedom ($\nu$) is also dependent on $\tau$:\\

\begin{equation}
\begin{split}
&G(\tau | \Sigma,\alpha^*, \beta^*) \propto \frac{|\Psi|^\frac{c\tau}{2} |\Sigma|^{-\frac{c\tau}{2}}}{2^{\frac{c\tau p}{2}} \Gamma_p(\frac{c\tau}{2})}  \tau ^ {\alpha - 1} \exp\left( - \tau\ \left( \frac{1}{2} \text{tr}(\Psi \Sigma^{-1}) + \beta \right)  \right) \\
\end{split}
\end{equation}

To conclude, the Inverse-Wishart is only conjugate to the Gamma for the scale parameter (or a scaling coefficient over parameter $\Psi$), and cannot be used as a conjugate prior for deriving the posterior distribution of $\tau_\Sigma$. 


\section{Appendix C: Conjugacy for $\tau_\mu$}

Let us consider a multivariate Normal distribution in a fully general case (Eq. \ref{eq:GammConj}),  with a conjugate Gamma prior over random variable $\tau$. We will show that the conjugacy holds for this setting, through expanding the right hand side of the proportionality and deriving the posterior hyper-parameters in the presence of a {\it single} sample of data ($A$), i.e. $N=1$. The resulting parameters can be easily extended to generalise to the case of $N$ observations.

\begin{equation}
\begin{split}
&G(\tau | A, \alpha^*, \beta^*) \propto N(A | \mu, \frac{\Sigma}{\tau \sigma}) G(\tau | \alpha, \beta), \hspace{1cm} \sigma, \tau > 0\\
&\propto \frac{1}{\sqrt{(2\pi)^2 \frac{|\Sigma|}{\tau \sigma}}} \exp(- \frac{\tau \sigma}{2} (A - \mu)^T \Sigma ^{-1} (A - \mu))) \times \frac{\beta ^ \alpha}{\Gamma(\alpha)} \tau^{\alpha - 1} \exp(- \beta \tau)\\
&\text{discarding the terms that are independent of the random variable $\tau$, we will have:}\\
&\propto \tau^{1/2} . \tau^{\alpha - 1} \exp(- \beta \tau - \frac{\tau \sigma}{2} (A - \mu)^T \Sigma ^{-1} (A - \mu))) \\
&\propto \tau^{\alpha - 1/2} \exp(- \tau (\beta + \frac{\sigma}{2} (A - \mu)^T \Sigma ^{-1} (A - \mu)))) \\
\end{split}
\label{eq:GammConj}
\end{equation}

The remaining terms are proportional to a Gamma distribution with the following parameters:

\begin{equation}
\begin{split}
\alpha^* = \alpha + 1/2 , \hspace{1cm} \beta^* = \beta + \frac{\sigma}{2} (A - \mu)^T \Sigma ^{-1} (A - \mu) \\
\end{split}
\end{equation}


\section {Appendix D: Impacts of $\tau$ on parameter distributions}

In this appendix, we explore the impact of $\tau$ on changing the mean and covariance of the Inverse-Wishart.
As mentioned in the paper, approximately in all cases the mean stays unchanged and the variance is scaled inversely to the learning rate.

\begin{equation}
\label{power}
\begin{split}
&IW(\Sigma|\Psi,\nu)^\tau \propto \left(|\Sigma|^{-\frac{\nu+p+1}{2}} \exp\left( {-\frac{1}{2}} tr(\Psi \Sigma^{-1}) \right)\right)^\tau \\ \\
&\left(|\Sigma|^{-\frac{\nu+p+1}{2}} \exp\left( {-\frac{1}{2}} tr(\Psi \Sigma^{-1}) \right)\right)^\tau = |\Sigma|^{-\frac{\tau(\nu+p+1)}{2}} \exp\left( {-\frac{1}{2}} tr(\tau\Psi \Sigma^{-1}) \right) \\
&\hspace{5.5cm} \approx |\Sigma|^{-\frac{\tau\nu+p+1}{2}} \exp\left( {-\frac{1}{2}} tr(\tau\Psi \Sigma^{-1}) \right)  \\ \\
&\Rightarrow \hspace{0.5cm} IW(\Sigma|\Psi,\nu)^\tau \propto IW(\Sigma|\tau\Psi,\tau\nu) \\
\end{split}
\end{equation}

Accepting the approximation above, the resulting $\Sigma$ samples are drawn approximately around the same mean, but with a scaled variance. When $0 \leq \tau_\Sigma < 1$ the variance increases, whereas for $\tau_\Sigma > 1$ the distribution is more peaky. In other words, the posterior samples of $\Sigma$ in the former case are allowed to move away from the IW mean, tending to have greater adaptability towards the current observed data, but in the latter case the posterior samples concentrate around the prior mean, discouraging covariance adaptation.

\begin{equation}
\begin{split}
&\text{mean of } \Sigma \sim IW(\Psi, \nu): \hspace{2cm}  M_\Sigma = \frac{\Psi}{\nu + p + 1}, \\
&\text{mean of } \Sigma^{\small(\tau)} \sim IW(\tau \Psi, \tau \nu): \hspace{1.2cm}  M_\Sigma^{\small(\tau)} = \frac{\tau \Psi}{\tau \nu + p + 1} \approx M_\Sigma \\
&\text{variance of }\Sigma \sim IW(\Psi, \nu): \hspace{1.6cm} V_\Sigma \approx \frac{\Psi_{ij}^2}{\nu^3}\\
&\text{variance of }\Sigma^{\small(\tau)} \sim IW(\tau \Psi, \tau \nu): \hspace{0.8cm} V_\Sigma^{\small(\tau)} \approx \frac{\tau^2 \Psi_{ij}^2}{\tau^3 \nu^3} \approx V_\Sigma/ \tau
\end{split}
\end{equation}



\bibliographystyle{unsrt}
\bibliography{misc}
\end{document}